\documentclass[]{article} 
\usepackage{tiaStyleFile}

\usepackage{bm}
\usepackage{amsthm}
\usepackage[toc,symbols]{glossaries}
\usepackage{makecell} 
\usepackage{pifont}
\usepackage{epigraph}

\usepackage[numbers]{natbib}

\usetikzlibrary{calc}

\usepackage{xcolor}
\definecolor{python1}{HTML}{1f77b4}
\definecolor{python2}{HTML}{ff7f0e}
\definecolor{python3}{HTML}{2ca02c}
\definecolor{python4}{HTML}{d62728}
\definecolor{python5}{HTML}{9467bd}
\definecolor{python6}{HTML}{8c564b}
\definecolor{python7}{HTML}{e377c2}
\definecolor{python8}{HTML}{7f7f7f}
\definecolor{python9}{HTML}{bcbd22}
\definecolor{python10}{HTML}{17becf}
\definecolor{designcolor}{HTML}{1f77b4}

\newcommand{\green}[1]{\textcolor{python3}{#1}}

\usepackage{hyperref}
\newtheorem{thm}{Theorem}[section]

\newcommand\X{\ding{55}}

\graphicspath{ {img/} }

\title{
Good Things Come in Pairs: \\ Paired Autoencoders for Inverse Problems}

\author{Matthias Chung, Emory University, USA\footnote{\url{mailto:matthias.chung@emory.edu}}, Bas Peters, Earth Dynamics AI, Canada\footnote{\url{mailto:bas@earthdynamics.ai}}, and Jack Michael Solomon, Emory University, USA\footnote{\url{mailto:jack.michael.solomon@emory.edu}}}

\begin{document}

\maketitle

\section*{Abstract}
 In this work, we discuss recent advances in data-driven approaches for inverse problems. In particular, we focus on the \emph{paired autoencoder} framework, which has proven to be a powerful tool for solving inverse problems in scientific computing. The paired autoencoder framework is a novel approach that leverages the strengths of both data-driven and model-based methods by projecting both the data and the quantity of interest into a latent space and mapping these latent spaces to provide surrogate forward and inverse mappings. We illustrate the advantages of this approach through numerical experiments, including seismic imaging and classical inpainting: nonlinear and linear inverse problems, respectively. Although the paired autoencoder framework is likelihood-free, it generates multiple data- and model-based reconstruction metrics that help assess whether examples are in or out-of-distribution. In addition to direct model estimates from data, the paired autoencoder enables latent-space refinement to fit the observed data accurately. Numerical experiments show that this procedure, combined with the latent-space initial guess, is essential for high-quality estimates, even when data noise exceeds the training regime. We also introduce two novel variants that combine variational and paired autoencoder ideas, maintaining the original benefits while enabling sampling for uncertainty analysis.

\paragraph{Keywords:} Inverse Problem, Autoencoder, Data-Driven, Model-Aware, Latent Space, Out-Of-Distribution, Likelihood-Free, Variational Autoencoder

\section{Introduction}\label{sec1}

This work discusses a novel class of data-driven methodologies for tackling inverse problems, leveraging the power of paired autoencoders. This framework centers on projecting both the observed data and the corresponding quantities of interest into their respective lower-dimensional latent spaces. By learning the intricate mappings between these latent representations, we effectively construct a surrogate model capable of approximating both the forward and inverse processes. This framework demonstrates significant performance gains compared to traditional data-driven approaches, yielding high-quality estimates for the recovered quantities of interest. In contrast to traditional data-driven machine learning approaches for inverse problems, paired autoencoders offer statistical measures to assess the reliability of the inverse solution, therefore addressing a key limitation of detecting out-of-distribution data. Moreover, variational-based paired autoencoder approaches extend this capability by providing uncertainty estimates for the inverse solution, as demonstrated through various numerical experiments.

The remainder of this work is structured to guide the reader through the intricacies of the paired autoencoder methodology. We begin by providing a comprehensive introduction to the realm of inverse problems in Section~\ref{sec:ip}, alongside a discussion of the classical techniques commonly employed to address them. Recognizing the inherent challenges associated with these traditional methods, we then present a concise overview of recent advancements in data-driven approaches in Section~\ref{sec:datadrivenip}, highlighting their potential to circumvent these limitations.

Before delving into the specifics of our paired autoencoder approach, we dedicate Section~\ref{subsection:autoencoders} to a thorough exploration of its fundamental building block: the autoencoder. We start by discussing the elementary principles of autoencoders and their theoretical underpinnings, including key results derived from the analysis of linear autoencoders in Section~\ref{subsection:autoencoder}. We then elaborate on the concept of latent space representations in Section~\ref{subsection:latent}, demonstrating how autoencoders can effectively capture the essential features of complex data and models in a compact form. Furthermore, we explore the variational variants of autoencoders in Section~\ref{subsection:vae}, which introduce a probabilistic perspective to latent space learning.

With this foundational knowledge in place, we proceed to detail the paired autoencoder framework. We first describe its general setup and architecture in Section~\ref{subsection:setup}, followed by a discussion on how to quantitatively assess the quality of reconstructions achieved through this method in Section~\ref{subsection:quality}. Subsequently, we provide a comprehensive explanation of the construction and training process for paired autoencoders in Section~\ref{subsection:lsinf}.

In Section~\ref{sec:vpae}, we explore extensions of the paired autoencoder framework aimed at incorporating uncertainty estimation using a variational paired autoencoder. Specifically, we examine two methodologies, including the use of a variational autoencoder as outlined in Section~\ref{subsection:vpae}, its corresponding inference mechanism discussed in Section~\ref{subsection:vpaeinf}, and the integration of probabilistic modeling in the latent space via a variational encoder-decoder, as presented in Section~\ref{subsection:vpaelatent}.

To underscore the practical utility and advantages of paired autoencoders, we showcase their application to two distinct prototype problems. The first application focuses on simple image restoration tasks in Section~\ref{subsection:imagerestore}, demonstrating the framework's ability to recover clean images from noisy or degraded observations. The second application tackles a more complex problem in seismic inference in Section~\ref{subsection:seismic}, highlighting the potential of paired autoencoders for extracting valuable subsurface information from seismic data. The network architectures and implementation details are presented in Appendix~\ref{appendix}. Finally, we conclude this work with a comprehensive discussion of our findings and outline promising future research directions within the burgeoning field of paired autoencoders in Section~\ref{sec:discuss}.

\subsection{Inverse Problems}\label{sec:ip}

Many scientific and engineering endeavors involve understanding a system's properties through indirect measurements. This process often leads to what are known as \emph{inverse problems}. Unlike direct or forward problems, where causes are known and effects are predicted, inverse problems aim to infer the causes from observed effects, \cite{hansen2010discrete}.

In mathematical terms, let us consider a system described by a mathematical operator $F\colon\mathcal{X} \to \mathcal{Y}$ that maps quantities of interest (model parameters or controls) $x \in \mathcal{X}$ either discrete or continuous to observations $y \in \mathcal{Y}$ potentially contaminated with some additive unbiased noise $\epsilon \in \mathcal{Y}$. The forward operator $F = P\circ u$ constitutes of the solution of the governing equations $u\colon \mathcal{X} \to \mathcal{U}$ typically described by a partial or ordinary differential equation (PDE/ODE), which is assumed to be fully determined by its model parameters $x$, and a projection $P\colon \mathcal{U} \to \mathcal{Y}$ onto the observation space $\mathcal{Y}$, i.e.,
$$
(P\circ u) (x) + \epsilon = y.
$$

Given the solution of the governing equations $u$, the projection $P$, the observations $y$, and some assumptions about the noise, the inverse problem aims to reconstruct the unknown model parameters $x$, frequently termed as \emph{quantity of interest (QoI)} \cite{hansen2010discrete}. The \emph{variational inverse problem} may be stated as an optimization problem of the form 
\begin{equation}\label{eq:VariationalInverse}
    \min_{x \in \mathcal{X}} \ \mathcal{L}(y, F(x)) + \mathcal{R}(x), 
\end{equation}
where $\mathcal{L}$ denotes the data misfit and $\mathcal{R}$ a regularization term. The regularization encodes prior knowledge about the QoIs, e.g., smoothness, sparsity, or positivity \cite{VogelInverse}. Then, a solution to the inverse problem is the QoI $\hat x$ that minimizes the objective function.

Bayesian inference provides a probabilistic framework for inverse problems. It involves determining the posterior distribution of the unknown QoI $x$ given the observed data $y$, i.e.,
  $$
  \pi_{\text{post}}(x \mid y) \propto \pi_{\text{like}}(y\mid x) \cdot \pi_{\text{prior}}(x)
  $$
where $\pi_{\text{prior}}$ is the prior belief about the QoI, $\pi_{\text{like}}$ is the likelihood of observing the data given the QoI, and $\pi_{\text{post}}$ is the posterior distribution \cite{calvetti2007introduction}. While classical inversion techniques are powerful tools, they often face challenges with large-scale problems, ill-posedness, and computationally intensive forward simulations. 

Inverse problems are numerous. For instance, in medical imaging, they are used to reconstruct images of internal organs from measurements obtained through techniques like X-ray computed tomography (CT) or magnetic resonance imaging (MRI) \cite{bertero2021introduction}, and others. Geophysics employs inverse methods to determine the Earth's subsurface structure by analyzing seismic \cite{TarantolaA,PrattFWI} and electromagnetic \cite{haber2014computational} data. In non-destructive testing, these techniques are crucial for identifying defects within materials using ultrasonic or electromagnetic measurements \cite{uhlmann2014inverse}. Remote sensing relies heavily on inverse problem-solving to retrieve atmospheric and surface properties from satellite observations \cite{doicu2010numerical}. More broadly in image processing, tasks like inpainting -- the reconstruction of missing or damaged image regions -- require inferring missing pixel values based on the surrounding image content \cite{guillemot2013image}.

A key challenge in inverse problems is their ill-posedness, meaning solutions may not exist (no model perfectly explains the data), may not be unique (multiple models fit the data equally well), or may be unstable (small data perturbations cause large changes in the estimated model), \cite{Hadamard}. Consequently, specialized techniques are essential. These include regularization, which introduces constraints or prior information to stabilize the solution; optimization, which frames the problem as {minimizing a loss between observations and forward modeled data}; and statistical methods, which incorporate statistical models of data and unknowns to quantify uncertainty \cite{calvetti2007introduction}. The study of inverse problems is a vibrant and interdisciplinary field, drawing upon mathematics, physics, statistics, and computer science. It plays a crucial role in advancing our understanding of complex systems and enabling new technologies across various domains. Limitations of classical inversion techniques are evident in applications, where large observational errors and simplified mathematical models can lead to local minima or constitute multi-modal posterior distributions. Consequently, classical inversion techniques become impractical for such applications without additional processing. Advanced techniques are necessary to overcome these challenges.

\subsection{Machine Learning Approaches for Inverse Problems}\label{sec:datadrivenip}
The emergence of data-driven approaches, particularly deep neural networks (DNNs), has significantly impacted the field of inverse problems \cite{jin2017deep,de2005learning, arridge2019solving, NNregularizationReview,mukherjee2023learned}. Traditionally, solving inverse problems relied heavily on analytical methods, numerical simulations, optimization algorithms, and sampling methods. Data-driven approaches aim to circumvent the challenges by utilizing those inversion methodologies. Below, we provide a brief overview of some deep learning approaches for inverse problems, to highlight some connections and differences to paired autoencoders. {Furthermore, Table ~\ref{tab:MethodsOverview} provides a summary of applicability of the various approaches, based on the availability of a forward operator, observed data, and model-data pairs during training and inference.}

\paragraph*{End-to-End methods.} For instance, consider learning the mapping between measurements $y$ and model parameters $x$ directly from training data. By assuming a true underlying mapping $\Psi\colon\mathcal{Y} \to \mathcal{X}$, training a neural network on a large dataset of paired input-target samples (i.e., pairs of measurements $y$ and corresponding QoI $x$) may learn to approximate the inverse mapping, effectively bypassing the need for explicit inversion algorithms. To be more precise, these techniques train neural networks, such as $\Phi: \mathcal{Y}\times \mathcal{T} \to \mathcal{X}$, to directly map observed data $y$ onto estimates $\hat x =\Phi(y;\theta)$, where $\theta \in \mathcal{T}$ are trainable network parameters. To improve readability, we often omit neural networks' explicit network parameter dependence, for instance, denoting them simply as $\Phi (\,\cdot\,)$. {In the supervised learning scenario we sample pairs of measurements $y$ and corresponding QoI $x$ from the joint distribution $(x,y) \sim \pi$. Denote the input and target marginal distributions using $\pi_y$ and $\pi_x$. Given sufficient and representative training pairs $(x_i,y_i)_{i = 1}^N$ from the joint distribution}, these end-to-end approaches have shown impressive results in producing reasonable reconstructions without explicit reliance on physical models. Because there is no unique, one-to-one relation between $x$ and $y$ for many inverse problems, the trained network implicitly contains implicit regularization: Given an input $y$, a non-stochastic network returns a single $\hat x$ out of all $x$ that generate $y$ under the forward operator $F$. 

A subset of the end-to-end methods is inspired by algorithms for optimizing \eqref{eq:VariationalInverse}, so-called unrolled optimization methods \cite{lsp,8683124,monga2021algorithm,le2024unfolded}. Unrolled optimization networks define a fixed number of iterations of a selected optimization algorithm. For each iteration, some of the components of the iterative algorithm are defined as learnable. For instance, if we want to stay close to the original optimization algorithm, we can make only the step size/learning rate a learnable parameter. In case of steepest descent applied to \eqref{eq:VariationalInverse}, we obtain the iteration
\begin{equation}\label{eq:Unrolled1}
x_{k+1} = x_k - \theta_k \nabla_{x_k}  \big( \mathcal{L}(y, F(x)) + \mathcal{R}(x) \big), \quad k=1,\ldots,k_\text{max},
\end{equation}
where $k$ is the iteration counter, we learn a single step-size parameter per iteration. Training proceeds via
{
\begin{equation}
    \min_\theta  \ \mathbb{E}_{(x,y) \sim \pi} \ \| x - \hat{x}_{k_\text{max}}(y; \theta) \|^2_2,
\end{equation}
}
where $\hat{x}_{k_\text{max}}(y; \theta)$ is the output of the unrolled algorithm \eqref{eq:Unrolled1}, which depends on the input data $y$ and learnable parameters. Each training iteration requires pairs $(x_i,y_i)_{j = 1}^J$ of data and true QoI. Other versions of unrolled algorithms may (partially) learn the forward operator by unrolling the optimization algorithm as
\begin{equation}\label{eq:Unrolled2}
x_{k+1} = x_k - \alpha \nabla_{x_k}  \big( \mathcal{L}(y, \Phi(x; \theta)) + \mathcal{R}(x) \big), \quad k=1, \ldots,k_\text{max},
\end{equation}
where $\Phi(x; \theta)$ replaces or augments the forward operator and now depends on learnable parameters $\theta$. Another unrolling approach is using a learnable projection or proximal operator $P_{\theta_k}$ via
\begin{equation}\label{eq:Unrolled3}
x_{k+1} = P_{\theta_k} \left( x_k - \alpha \nabla_{x_k} \mathcal{L}(y, F(x)) \right), \quad k=1,\ldots,k_\text{max},
\end{equation}
where the regularization penalty was removed, and all prior knowledge is now learnable as part of the iteration-dependent $P_{\theta_k}$. The learnable operator may range from a single convolution to a neural network with a modest number of layers. In general, the learned  $P_{\theta_k}$ only resembles a proximal operator. Learning a true proximal operator requires specialized designs and training, see, e.g., \cite{fangs}. This data-driven approach has shown promising results in various applications, including image reconstruction, medical imaging, and seismic inversion, offering faster and potentially more accurate solutions than traditional data-driven and model-based approaches, \cite{monga2021algorithm}. 

End-to-end methods share the requirement of accessing pairs of data $y$ and models $x$ at training time. Sometimes there is a lack of available data $y$, or it is computationally too costly to generate $y$ corresponding to a large number of $x$. In these scenarios, we may use other approaches that only utilize a collection of $x$ examples for learning prior knowledge, as discussed below.

\paragraph*{Prior Learning.}

Here, when we refer to prior learning, we mean learning about the QoI using data samples. This is different from how prior information is used in Bayesian inverse problems (through prior distributions). Prior learning methods differ from previously discussed approaches since they can learn directly from the samples $x$.

A first step to learning prior information is replacing or augmenting the regularization functional $\mathcal{R}$ by a trainable function $\Phi(x;\theta)$ \cite{afkham2021learning}, or a relatively low number of parameters of a hand-crafted regularization functional \cite{haber2003learning} in
\begin{equation}\label{eq:LearnedVariationalInverse}
    \min_{x \in \mathcal{X}} \ \mathcal{L}(y, F(x)) + \Phi(x; \theta). 
\end{equation}
Upon the availability of larger training datasets and more powerful computational resources, we may aim to learn more from the data by training (deep) neural networks (DNN) with many parameters. Various approaches exist to learn a regularizer $\mathcal{R}$ as a deep neural network. In \cite{NEURIPS2018_d903e960} the authors train the regularizer using {(unpaired)} examples images $x \sim \pi_x$ from the distribution of ground truth images, and images $x_c \sim \pi_c$ from the distribution of corrupted images. The idea is that a good regularization function assumes a large value for corrupted images, and a low value for correct images. Note that corrupted images are often generated by degrading good images, i.e., samples $x_c$ from $\pi_c$ are conditioned on $x$. The corresponding minimization problem reads
\begin{equation}\label{eq:AdvReg}
    \min_\theta \ \mathbb{E}_{x \sim \pi_x}  \Phi(x; \theta) - \mathbb{E}_{x_c \sim \pi_c} \Phi(x_c; \theta),
\end{equation}
which is a simplified version of \cite{NEURIPS2018_d903e960}. See \cite{alberti2021learning, kobler2020total, mukherjee2020learned,10223264} for other options to train a regularizer. Once trained, the regularizer is fixed and used in \eqref{eq:LearnedVariationalInverse}. The objective in Equation~\ref{eq:AdvReg} is conceptually somewhat similar to maximum likelihood learning Energy-Based Models (EBMs, \cite{ContrastiveBP,NEURIPS2019_378a063b}) via gradients of the energy functionals. See \cite{zach2021computed, balcerak2025energy} for examples of applying EBMs to inverse problems.

{An area of active research concerns Flow-Matching (FM) \cite{albergobuilding, lipman2022flow} for generative modeling, where a learned space and time-dependent velocity field $v(x,t;\theta)$ maps a simple source distribution (often Gaussian) into the target data distribution. The likelihood of $x$ is computable using the trained flow \cite{lipman2022flow}, and may be used as the prior in Equation~\ref{eq:LearnedVariationalInverse}. The regularizer then reads \cite{zhang2024flow}

\begin{equation}
 \Phi(x; \theta)_\text{flow} = \| x_{t=0}\|_2^2 + \int_{t=0}^{t=1} \operatorname{tr}\bigg( \frac{\partial}{\partial x} v(x,t;\theta) \bigg) dt,
\end{equation}

which involves integrating an ODE back in time, starting at the current QoI estimate $x$, while computing the trace \cite{grathwohl2018ffjord}. Here, the flow end-time $t=1$ corresponds to the target distribution, and $t=0$ corresponds to the source. The trace computation (approximation) and time-integration are typically considered computationally challenging \cite{song2021maximum, feng2023score}; features that are shared by both diffusion models and flow matching.}

The learned regularizer $\Phi(x, \theta)$ serves as a critic or discriminator of the input $x$. This point of view bridges this discussion to generative models like Generative Adversarial Networks (GANs) \cite{goodfellow2014generative} and Variational Auto Encoders (VAEs) \cite{kingma2022autoencodingvariationalbayes}. Both GANs and VAEs learn a decoder (generator) $d(z): \mathcal{Z} \rightarrow \mathcal{X}$ that maps samples from a latent source distribution $\pi_z$ (typically Gaussian noise) to samples from the true image distribution $\pi_x$. Unlike regularization functionals that output a scalar value, a decoder maps $\mathcal{Z} \rightarrow \mathcal{X}$. If a generator outputs only good images, from $\mathcal{X}$, we may optimize not over the (physical) parameters $x$, but over the latent variables $z$ as follows:
\begin{equation}\label{eq:ParameterizedInversion}
    \min_{z} \mathcal{L}(y, F(d(z)).
\end{equation}
This is now an unconstrained problem without any penalty terms, where all information learned from examples is included in $d(z)$. If the original problem is non-convex, the new problem remains non-convex, but practical examples show that gradient-based methods may achieve lower function values.

Another important class of methods learns a direct map from corrupted, incomplete, or otherwise unrealistic images to ground truth images. This is reminiscent of the function of proximal and projection operators in the context of imaging inverse problems. Denote this operator again by $\Phi(x; \theta)$, but this time the network provides a correction of the input instead of a direct model estimate. A straightforward training objective is
{
\begin{equation}
    \min_\theta \mathbb{E}_{(x_c, x) \sim \pi} \| \Phi(x_c; \theta) - x \|,
\end{equation}
}
where we sample an example image $x$, corrupt it, and measure how well $\Phi(x_c; \theta)$ removes the corruption from $x_c$. Once trained, the operator is deployed in any proximal optimization algorithm, the simplest instance being the proximal-gradient method
\begin{equation}\label{eq:PnP}
    x_{k+1} = \Phi \left( x_k - \alpha \nabla_x \mathcal{L}(y, F(x)) ; \theta\right).
\end{equation}
Plug-and-play methods of the above form and its variants \cite{10004791, 8068267, 10328845,10274489,PnPFWI1, zheng2025inversebench} often separate the solution of the inverse problem and learning from prior knowledge, while building on the observation that black-box denoisers are effective if used instead of actual proximal operators \cite{venkatakrishnan2013plug}. Beyond empirical efficacy, convergence results are known for \eqref{eq:PnP} given certain contexts such as strictly convex $F(x)$ or non-expansiveness of $\Phi(x; \theta)$ \cite{ryu2019plug,9380942,QN_PnP,hauptmann2024convergent, Terris_2024_CVPR}.

{\paragraph*{Learning Data-Dependent Flows, Conditional \& Guided generation.}
Flow-matching (FM, \cite{albergobuilding, lipman2022flow}) recently gained popularity in the context of solving inverse problems. One of the reasons behind this is the various ways in which a flow trained via FM may be utilized. Here, we informally describe a few, while omitting some details. Some of the concepts in this paragraph also apply to diffusion models, see, e.g., \cite{pokle2024trainingfree} and \cite[Chapter.~10]{lipman2024flow} for an overview of the connection between the methods. 

Consider a trainable velocity field $v(x,t; \theta)$ with parameters $\theta$, space (data) coordinates $x$ and time $t \in [0,1]$. The goal is to learn how samples from the source distribution, $\pi_s$, flow to the target distribution $\pi_x$. The source distribution is often chosen as a simple Gaussian, but can also be the distribution of observed data or corrupted images-- which are of interest in the inverse problem context. The target distribution represents the QoI: physical model parameters or true images, for instance.

Before illustrating how to use FM for inverse problems, consider the specific training objective \cite{liu2023flow,tong2024improving} (various other forms exist)
\begin{equation}\label{eq:FM}
\min_\theta \mathbb{E}_{t\sim U[0,1], x \sim \pi_x, y \sim \pi_y} \| v(x,t; \theta) - (x - y) \|_2^2, 
\end{equation}
where time $t$ is sampled uniformly $[0,1]$, and the $x$ and $y$ are sampled independently from each other, i.e, and unpaired and unsupervised setting. Integrating the velocity field in time (via the ODE $dx/dt = v(x,t; \theta)$ using e.g., forward Euler), starting with a sample $y \sim \pi_y$ will now generate a sample $x \sim \pi_x$. However, this does not mean that the generated sample from the QoI distribution will actually correspond to the sampled data $y$. 

Several extensions to the basic FM formulation in Equation~\ref{eq:FM} are more suitable to solve inverse problems. If paired samples $(x,y) \sim \pi$ are available, we can train the velocity via

\begin{equation}\label{eq:FMpaired}
\min_\theta \mathbb{E}_{t\sim U[0,1], (x,y) \sim \pi} \| v(x,t; \theta) - (x - y) \|_2^2, 
\end{equation}
which leads to promising results for image processing tasks like inpainting , super-resolution, and deblurring \cite{pmlr-v235-albergo24a}. 

An extension to both Equation~\ref{eq:FM} and Equation~\ref{eq:FMpaired} introduces a condition $c$, also referred to as a guidance signal \cite{ho2022cascaded, Saharia2023}. This condition may represent observed data, auxiliary information like an inpainting or data-acquisition mask, an image class-label, or even a textual description or caption. With additional conditions, Equation~\ref{eq:FMpaired} becomes

\begin{equation}\label{eq:FMpaired_conditional}
\min_\theta \mathbb{E}_{t\sim U[0,1], (x,y) \sim \pi(x,y|c)} \| v(x,t, c; \theta) - (x - y) \|_2^2, 
\end{equation}

which is similar to the formulation of \cite{pmlr-v235-albergo24a}, and where the paired data are sampled as $(x,y) \sim \pi(x,y|c)$, and the velocity field, $v(x,t, c; \theta)$, likely parameterized by a neural network, depends on $c$ as well. This technique also applies when the source distribution is chosen as Gaussian.

The above approaches require that the observed data $y$ and/or condition $c$ be available at the training time. When this is not the case, we may use techniques that include $y$ or $c$ at the inference stage. One such technique augments the velocity field with a guidance field $g(x,t,c)$ such that integrating the augmented field $v(x,t; \theta) + \alpha g(x,t,c)$ yields samples from the target distribution with desired properties, image-class, or adherence to observed data \cite{song2020score, daras2024survey, pokle2024trainingfree, feng2025on}. Because $g(x,t,c)$ may be deterministic, or learned separately, the terms require balancing via the scalar $\alpha>0$.

\paragraph*{Physics-Informed Neural Networks and Model-Constrained Learning.} In the last few years, Physics-Informed Neural Networks (PINNs) \cite{RAISSI2019686} have become a popular option for solving inverse problems in many applied domains.  Motivated in part by the high data demands of existing DNN-based approaches, PINNs introduce regularization terms to \emph{inform} the model of the physics guiding the system, thus driving the model towards physical solutions.  That is, prior information regarding the physics of a system is considered in model optimization. 

While PINNs are ubiquitous in numerical PDEs we focus on their use in inverse problems.  In particular, we consider supervised PDE parameter estimation problems and PDE-constrained inverse problems.  The former entails estimating the parameters of a PDE system from data, while the latter aims to solve an inverse problem where QoI are constrained by the physics of the system \cite{cuomo2022scientificmachinelearningphysicsinformed, toscano2024pinnspikansrecentadvances,NganyuTanyu_2023}.  
 
Assuming we have a forward differential operator, boundary conditions and spatial measurements $u(\chi_j)= \tau_j$ for $j = 1, \dots, m$, the solution of the PDE $u$ is modeled by a DNN, $\Phi(\,\cdot\,; \theta)$, parameterized by $\theta$. In the case of parameter estimation, these $\tau_j$ may coincide with our data $y_i$.  Training simultaneously learns an approximation for the solution of the PDE $u(\chi)$ by $\Phi(\chi; \theta)$, and our QoI through minimization of the following quantities (if applicable): first, the mismatch computation of the forward model at the estimated solution; second, the violation of the predicted solution of boundary conditions; third the violation of the predicted solution of the physics governing the system; lastly, the mismatch between $u$ and $\tau_j$ evaluated at each $\chi_j$. Derivatives of $\Phi(\chi; \theta)$ can be computed using automatic differentiation, and mean-squared-error (MSE) is a common choice to compute these metrics \cite{cuomo2022scientificmachinelearningphysicsinformed}.

PINNs have seen significant use in many pathological inverse problems.  For instance, \cite{JAGTAP2020113028, KHARAZMI2021113547, 10255379}, present variations of PINN-based approaches that are used to solve various PDE-discovery questions efficiently and accurately.  Furthermore, PINNs have seen successful application in geophysical inverse problems such as seismic inversion, and estimates of subsurface properties \cite{li2020coupled}.  PINNs have also been used to enhance inverse-problem solutions arising in medical-imaging modalities, see \cite{chen2023physics, banerjee2024pinnsmedicalimageanalysis, FATHI2020105729, fok2024deep, shone2023deep}. 

The main advantage of PINNs over other approaches is their flexibility in solving nonlinear, high-order inverse problems.   On the other hand, these methods require significant prior knowledge of the system, which may introduce additional uncertainty in the solution.  Additionally, PINNs suffer from the common challenges of deep learning-based approaches.  In particular, training becomes difficult for high-dimensional parameters and may require many iterations to converge \cite{lu2021physicsinformedneuralnetworkshard}.  This is particularly problematic when the evaluation of the PDE is expensive and required at each iteration \cite{cuomo2022scientificmachinelearningphysicsinformed}. Furthermore, PINNs may not generalize well.  A PINN trained for one problem may not generalize to a variation of the same problem \cite{nguyen2024taenmodelconstrainedtikhonovautoencoder}. 

\begin{table}[ht]
    \centering
    \resizebox{\textwidth}{!}{%
    \begin{tabular}{l|cccc}
       Method                       & \makecell{$F$ for \\ training} & \makecell{$F$ for \\ inference} & \makecell{Data $y$ \\ for training} & \makecell{Paired $\{x,y\}$ \\ for training}  \\ \hline
       End-to-end network              & \green{\X}       & \green{\X}       & \red{\checkmark} & \red{\checkmark}  \\
       Unrolled end-to-end             & \red{\checkmark} & \red{\checkmark} & \red{\checkmark} & \red{\checkmark}  \\
       Learn regularizer (penalty)     & \red{\checkmark} & \red{\checkmark} & \green{\X}       & \green{\X}        \\
       Learn generator (AE/VAE)        & \green{\X}       & \green{\X}       & \green{\X}       & \green{\X}        \\
       Learn generator (paired AE/VAE) & \green{\X}       & \green{\X}       & \red{\checkmark} & \red{\checkmark}  \\
       Learn Plug-and-Play             & \green{\X}       & \red{\checkmark} & \green{\X}       & \green{\X}        \\
       PINN                            & \red{\checkmark} & \green{\X}       & \red{\checkmark} & \red{\checkmark}  \\
       {Flow - paired/conditional sampling} & \green{\X} &  \green{\X}       & \red{\checkmark} &    \red{\checkmark} \\
       {Flow - controlled generation}       & \green{\X} &  \red{\checkmark} & \green{\X}       &    \green{\X}       \\
    \end{tabular}%
    }
    \caption{Methods overview in terms of favorable (green) and unfavorable properties. Here, the forward operator involves relatively costly ODE/PDE solves. The overview is general, and variants of methods, hybrids, and exceptions may deviate from this general overview.}
    \label{tab:MethodsOverview}
\end{table}

Model-constrained approaches are similar to PINNs, but different in that they parametrize the inverse map by a DNN.  Heuristically, these approaches learn the inverse mapping constrained by the forward map. Like PINNs, Model-constrained approaches attempt to ameliorate DNN data dependence by introducing additional information through regularization terms in optimization.  Assuming a forward map $F$ is available, the model is aware that the observations it sees come from the forward model. Letting $\Phi(y; \theta)$ denote the learned inversion, model training optimizes
\begin{align}
\min_\theta \bbE_{(x,y) \sim \pi}\norm[2]{x - \Phi(y; \theta)}^2 + \alpha\norm[2]{y - F(\Phi(y;\theta))}^2
\end{align}
for $\alpha > 0$ a regularization parameter. In some cases, the additional term can be interpreted correctly as a physics-informed regularization.

Several strategies exist to represent $\Phi$.  For example, \cite{goh2019solving} introduces the use of variational autoencoders and explores approximations of posterior distributions using this methodology.  Further data-efficient approaches have been proposed in which the inverse map learns the solution to the Tikhonov-regularized inverse problem \cite{nguyen2022tnetmodelconstrainedtikhonovnetwork, nguyen2024taenmodelconstrainedtikhonovautoencoder}.

\subsection{Autoencoders}\label{subsection:autoencoders}

Autoencoders are a class of neural networks $\Phi\colon \mathcal{X} \to \mathcal{X}$ designed to map inputs onto themselves. The autoencoder is typically decomposed into a \emph{composition} of two functions $\Phi(x) = (d \circ e)(x)$, with \emph{encoder} $e \colon \mathcal{X} \to \mathcal{Z}$ mapping into a latent space $\mathcal{Z}$ and a \emph{decoder} $d \colon \mathcal{Z} \to \mathcal{X}$ mapping a latent variable $z$ back onto the space $\mathcal{X}$. Note that the network parameters $\theta$ of the autoencoder can therefore be split into a set of network parameters for the encoder $\theta^e$ and decoder $\theta^d$, i.e., $\theta = (\theta^e, \theta^d)$. We denote the image of the encoder $\mathcal{Z}$ as the \emph{latent space}, and denote $z = e(x)$ as the latent representation of the input $x$. 
Given a smaller dimensionality of the latent space compared to the input space $\mathcal{X} \subset \mathcal{Z}$ is utilized in many applications leads to the autoencoder's distinctive hourglass or diabolo shape, illustrated in Figure~\ref{fig:autoencoder} and discussed in various papers such as \cite{schwenk1997training,li2023comprehensive,michelucci2022introduction, goodfellow2016deep}.  In contrast, overcomplete autoencoders, defined by $\mathcal{X} \supset \mathcal{Z}$, gained momentum in recent years, especially when combined with compressed sensing techniques in the efficient latent space representation \cite{chung2024sparse}. Besides compressing the latent space via a reduction in parameter count, or via sparsification of an overcomplete representation, there are also invertible neural networks \cite{ChangReversible,pmlr-v97-behrmann19a} that cannot change their number of parameters throughout the network. \cite{lensink2022fully} proposed a low-rank and sparse representation of the latent space for this case.

\begin{figure}
\centering
\begin{tikzpicture}
    	\node[fill=python1!50, minimum width=0.5cm, minimum height=2.5cm] (x) at (0,0) {$x$};
    	
    	\draw[fill=python9!50,draw=none] ([xshift=0.5cm]x.north east) -- ([xshift=2.5cm,yshift=0.5cm]x.east) -- ([xshift=2.5cm,yshift=-0.5cm]x.east) -- ([xshift=0.5cm]x.south east) -- cycle; 
    	\node at (1.75,0) {$\begin{matrix}\mbox{encoder} \\ e \end{matrix}$};
    	
    	\node[fill=python3!50, minimum width=0.5cm, minimum height=1.0cm] (Zx) at (3.5cm,0) {$z$};
    
     	\draw[fill=python4!50,draw=none] ([xshift=0.5cm]Zx.north east) -- ([xshift=2.5cm,yshift=0.75cm]Zx.north east) -- ([xshift=2.5cm,yshift=-0.75cm]Zx.south east) -- ([xshift=0.5cm]Zx.south east) -- cycle;
    	\node at (5.25,0) {$\begin{matrix}\mbox{decoder} \\ d \end{matrix}$};
    
    	\node[fill=python1!50, minimum width=0.5cm, minimum height=2.5cm] (B) at (7,0) {$x$}; 
    \end{tikzpicture}
    \caption{Schematic representation of an autoencoder neural network $\Phi$. The autoencoder consists of a composite mapping $\Phi = d \circ e$ where $e$ denotes the encoding and $d$ the decoding process. An encoded signal $z = e(x)$ is referred to as the latent variable or latent representation.}
    \label{fig:autoencoder}
\end{figure}
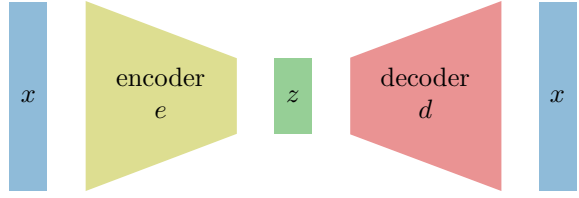   

The training of autoencoders is considered an \emph{unsupervised} or \emph{self-supervised} learning approach, where the network training learns to reconstruct the input data without relying on corresponding pre-defined labels or target data. The autoencoder's learning process involves minimizing a loss function between the original input and the reconstructed output. For instance, for the mean squared error (MSE) loss, we have
$$
 \min_{\theta} \ {\mathbb{E}_{x\sim \pi_x} \left\| d(e(x;\theta^e);\theta^d)-  x\right\|_2^2.}
$$

The encoding process $e$ is designed to retain only the `main features' of the input $x$; those characteristics that are most crucial for reconstructing the original input. Redundant or less significant details are discarded. Subsequently, the decoder component $d$ of the autoencoder attempts to reconstruct the original input from this compressed latent space representation.  By minimizing the loss, the network learns to effectively encode and decode the data, ensuring that the latent space representation accurately captures the input's most salient features.

\subsection{Linear Autoencoder} \label{subsection:autoencoder} In the case of linear autoencoders where both the encoder and decoder are linear transformations, i.e., $e = E\in \mathbb{R}^{r \times n}$ and $d = D\in \mathbb{R}^{n \times r}$ and each element of those matrices are trainable, with $r \leq n$ it is well known that the optimal solution is closely related to principal component analysis (PCA).  This relationship has been established by assuming {samples $x_1, \ldots, x_{n_x}$ drawn independently from the distribution $\pi_x$} with $n_x \gg r$ and the corresponding data matrix $[x_1,\ldots, x_{n_x}]$ has at least column rank $r$.  The mathematical equivalence between PCA and linear autoencoders has been established in several works, notably \cite{bourlard1988auto,baldi1989neural,plaut2018principal}. These studies demonstrate that the weight matrices of the encoder and decoder correspond to the eigenvectors of the data covariance matrix associated with the largest eigenvalues. This insight has been leveraged in various applications, including dimensionality reduction, denoising, and feature extraction, where linear autoencoders provide an interpretable and computationally efficient framework for data representation. For further details, see \cite{bao2020regularized}, which explores regularized extensions of linear autoencoders, and \cite{wang2014generalized}, which discusses their use in model reduction.

A deeper understanding of optimal linear autoencoders {$A = DE \in \mathbb{R}^{n \times n}$} may be gained by treating $x$ as a random variable\footnote{To simplify notation, we will use lowercase $x$ to represent a random variable, relying on context to distinguish between random variables and samples.}, and analyzing the problem through the lens of the \emph{Bayes risk} formulation, i.e.,
\begin{equation}\label{eq:bayesrisk}
  \min_{\text{rank}(A)\leq r} \ {\mathbb{E}_{x\sim \pi_x} } \left\| Ax - x\right\|_2^2.
\end{equation}
Here, ${\mathbb{E}_{x\sim \pi_x} }$ denotes the expectation of a random variable $x$. Let us assume that $x$ has a finite second moment that is symmetric positive definite  $\Gamma_x = L_x L_x^\top= \mathbb{E} \,x x^\top$ with a symmetric decomposition by a matrix $L_x \in \mathbb{R}^n$. In this case, the objective function in Equation~\ref{eq:bayesrisk} reduces to
\begin{align*} 
    {\mathbb{E}_{x\sim \pi_x} }  \left\| Ax - x\right\|_2^2 &= 
     \text{tr} \left( ( A - I) L_x L_x^\top( A - I)^\top \right) 
 =  \left\| AL_x - L_x \right\|_{\text{F}}^2,
\end{align*}
where  $\text{tr}(\,\cdot\,)$ denotes the trace of a matrix and  $\left\| \, \cdot \, \right\|_{\text{F}}$ denotes the Frobenius norm.

For $r = n$ the identity mapping $A = I$ is an optimal solution with arbitrary invertible matrix $E = K^{-1} \in \mathbb{R}^{n\times n}$ and $D = K$. For rank constraint problems $\ell < n$ an optimal low-rank solution can be found using a generalization of the Eckart–Young–Mirsky theorem \cite{friedland2007generalized, chung2017optimal}.
\begin{thm}[\cite{hart2025paired}] \label{theorem:fullrowrank}
    Let matrix $L_x \in \mathbb{R}^{n \times n}$ have full row rank with SVD given by $L_x = U \Sigma V^\top$. Then
    $$\hat A = U_{r} U_{r}^\top $$
    is a solution to the minimization problem
    $$\min_{\text{rank}(A)\leq r}  \left\|A L_x - L_x\right\|^2_{\text{F}} ,$$
    having a minimal Frobenius norm $\|\hat A\|_{\text{F}}= \sqrt{r}$ and $\| \hat A L_x - L_x\|^2_{\text{F}} = \sum_{k = r+1}^n \sigma_k(L_x)$, where $\sigma_k(L_x)$ denotes the $k$-th sorted singular value of $L$, $\sigma_k(L_x)\geq \sigma_{k+1}(L_x)$. This solution is unique if and only if either $r = n$
     or $1\leq r < n$ and $\sigma_r(L_x) > \sigma_{r+1}(L_x)$.
\end{thm} \label{theorem:aetheorem}

While the optimal autoencoder $\hat{A}$ may be uniquely determined, its individual encoder $\hat E$ and decoder $\hat D$ components are not. Specifically, for any invertible matrix $K \in \mathbb{R}^{r \times r}$, we can express the decoder as $\hat D = U_r K^{-1}$ and the encoder as $\hat E = K U_r^\top$, demonstrating the non-uniqueness.

\subsection{Latent Space Representation} \label{subsection:latent}
Let us consider an autoencoder $\Phi$ and train this autoencoder on MNIST training images $(x_i)_{i = 1}^{60,000}$ (not the corresponding labels), mapping images onto themselves. Assuming we are using just a three-dimensional latent space $\mathcal{Z} = \mathbb{R}^3$, we depict the encoded training samples $z_i = e(x_i)$ in Figure~\ref{fig:mnistlatent}, where we color-coded each sample according to its label. The latent space representation of the digits shows a clear separation between different labels, indicating that the autoencoder has learned to encode the essential (unsupervised) features of the data in the latent space. This property makes autoencoders useful for tasks such as dimensionality reduction, clustering, and anomaly detection, where the latent space representation can be used to identify patterns and similarities in the data.

\begin{figure}[h]
    \centering
    \includegraphics[width=0.45\textwidth]{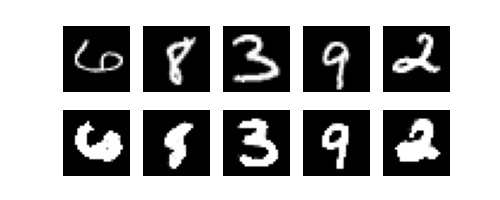}    
    \includegraphics[width=0.45\textwidth]{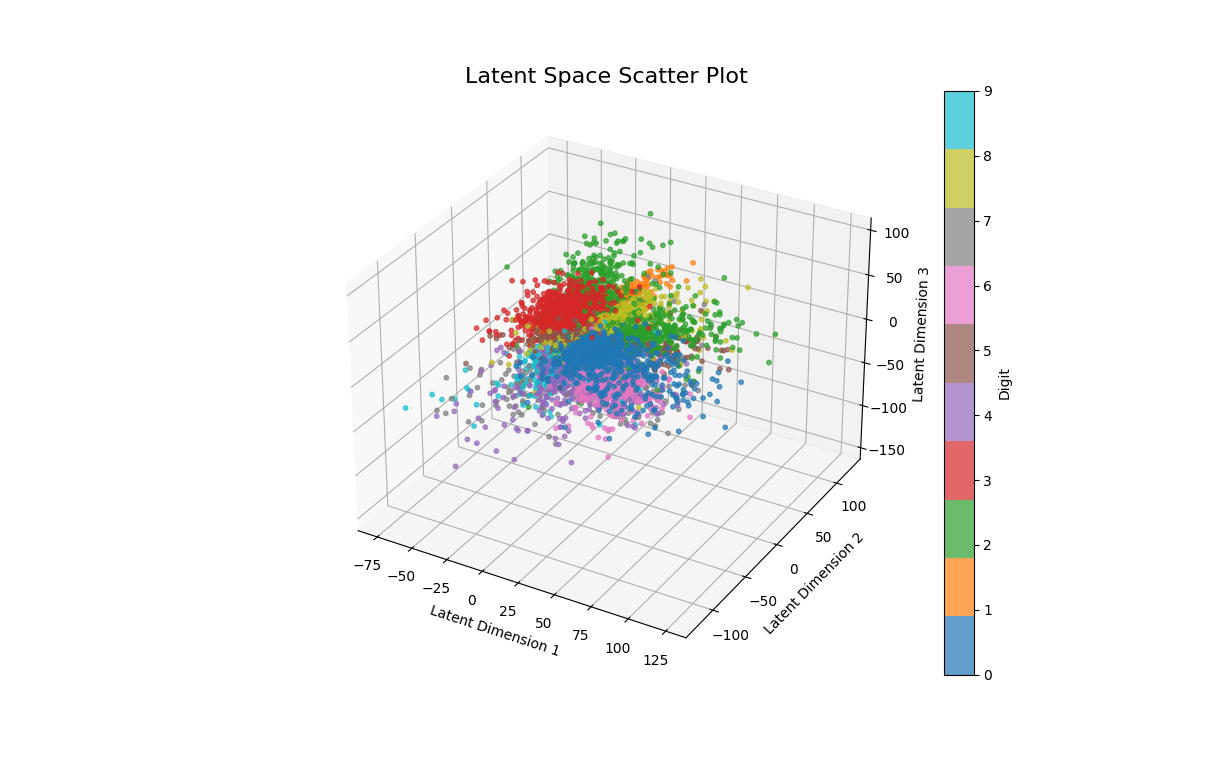}
    \caption{Latent space representation of MNIST digits encoded by an autoencoder. On the left, we present the original MNIST digit (at the top) alongside its reconstruction (below) obtained using a trained autoencoder sample $(d \circ e) (x_j) $ with a 3-dimensional latent space for various samples $x_j$. Correspondingly, on the right, we display a scatter plot of each sample's latent representation ($z_j = e(x_j)$), where the color of each point signifies the digit it represents (e.g., red for the digit ``three''). Despite the autoencoder not being provided with digit labels, the latent space representations naturally separate according to digit.}
    \label{fig:mnistlatent}
\end{figure}

As this example shows, data often exhibits low-dimensional structures that can be exploited to reduce its dimensionality with little loss of information. As demonstrated above, an autoencoder aims to identify these representative spaces. Similarly, mathematical models represent complex realities, and we can streamline model analysis by identifying the corresponding model parameters, which are typically far fewer than the dimensions of the model's behavior. For instance, consider dynamics generated by shallow water equations. The shallow water equations are a set of partial differential equations that describe the flow of a fluid layer where the horizontal length scale is significantly larger than the vertical length scale. These equations are fundamental in modeling various geophysical flows, including tsunamis, river flows, and atmospheric dynamics, by simplifying complex fluid behaviors into a manageable mathematical framework \cite{kinnmark2012shallow}. The time snapshots shown in Figure~\ref{fig:swe} (top) are both produced by the same model equations, only differing in the initial conditions Figure~\ref{fig:swe} (bottom). While physical models are often computationally demanding, we may want to identify a corresponding surrogate model that finds an appropriate low-dimensional representation of the model's behavior.

The latent space representation of the model parameters or QoI may be used to identify the underlying physical properties of the system, such as wave speed, amplitude, and frequency. This latent space representation can be used to analyze the model's behavior, identify important features, and make predictions about the system's dynamics.

\begin{figure}[h]
    \centering
    \includegraphics[width=0.25\textwidth]{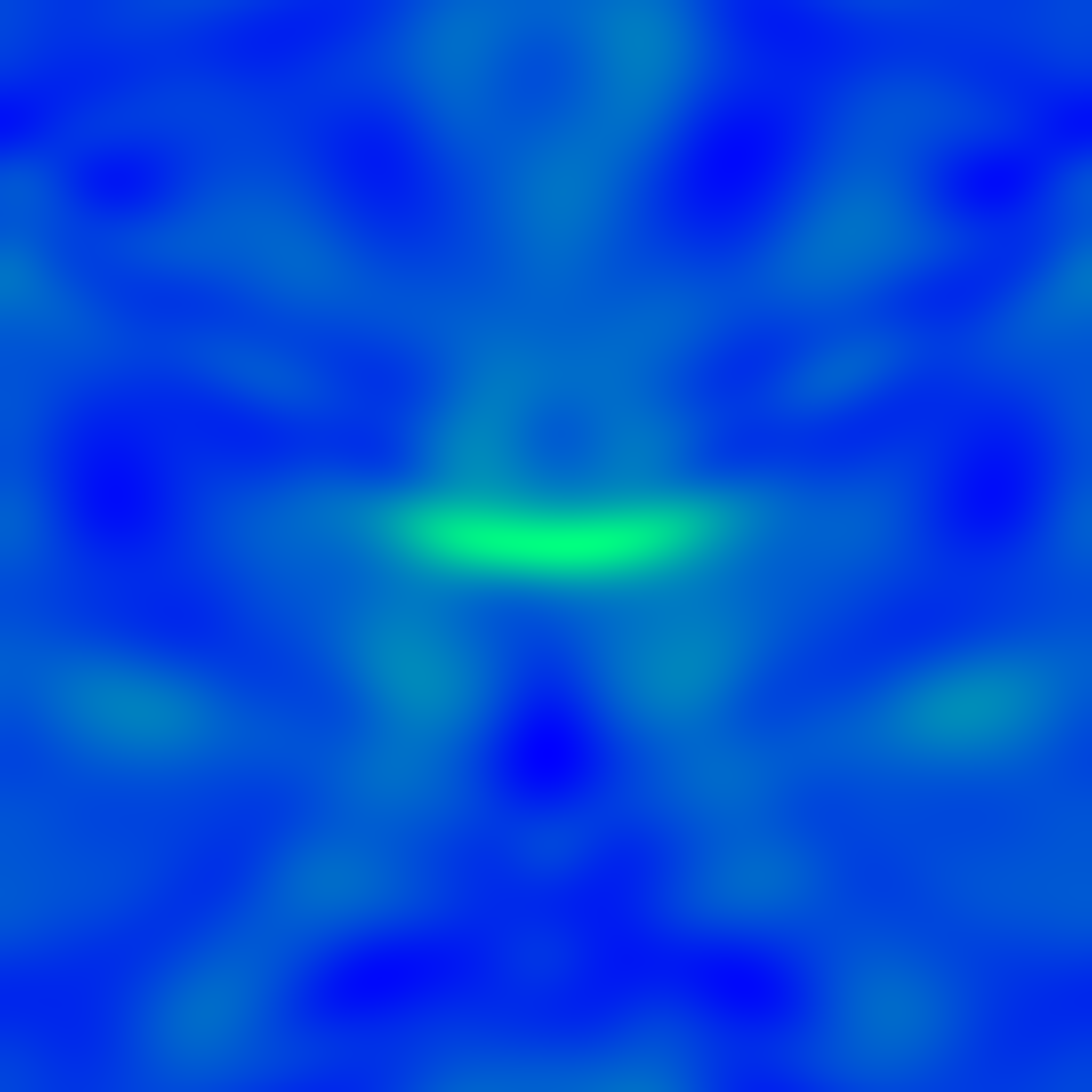}
    \includegraphics[width=0.25\textwidth]{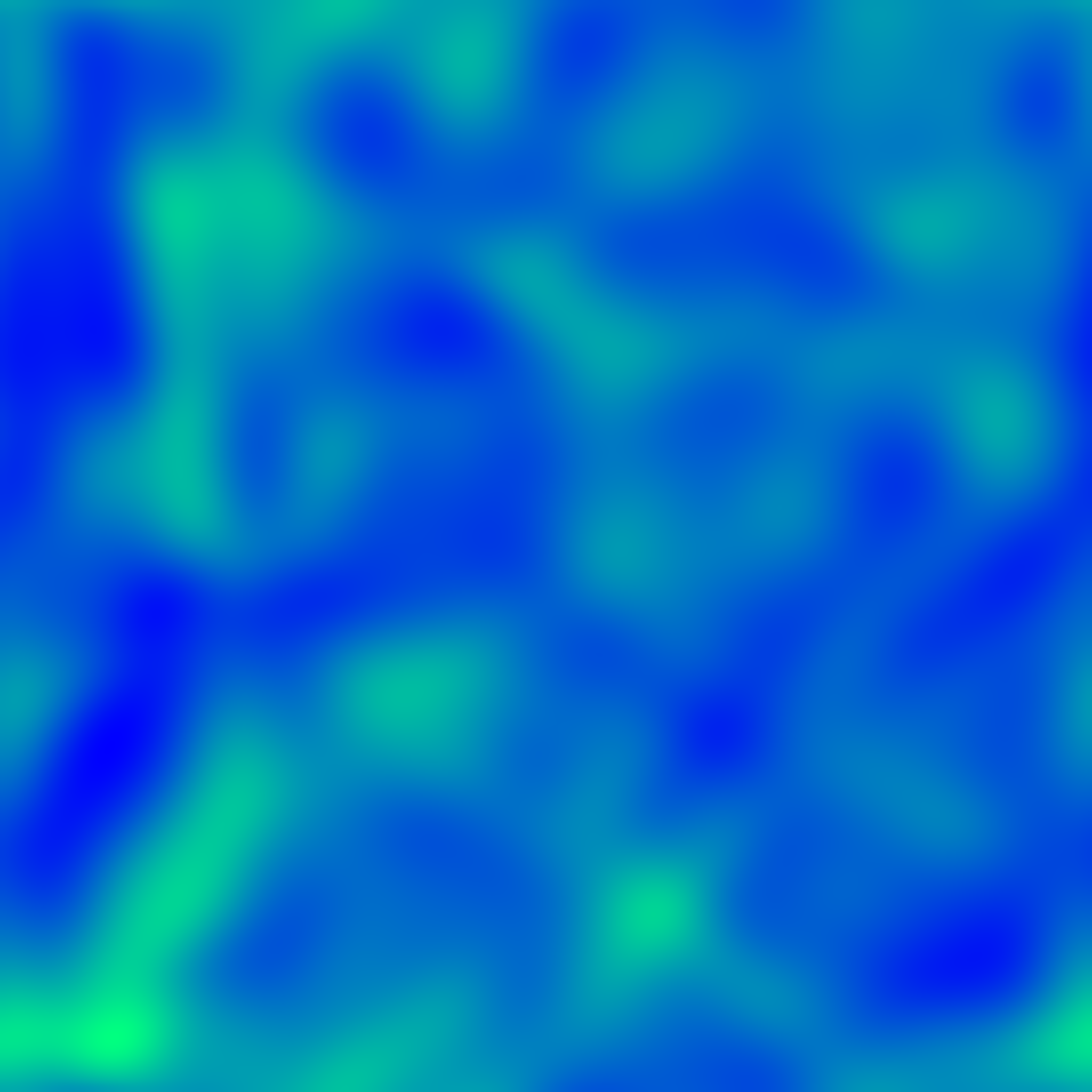}\\    
    \includegraphics[width=0.25\textwidth]{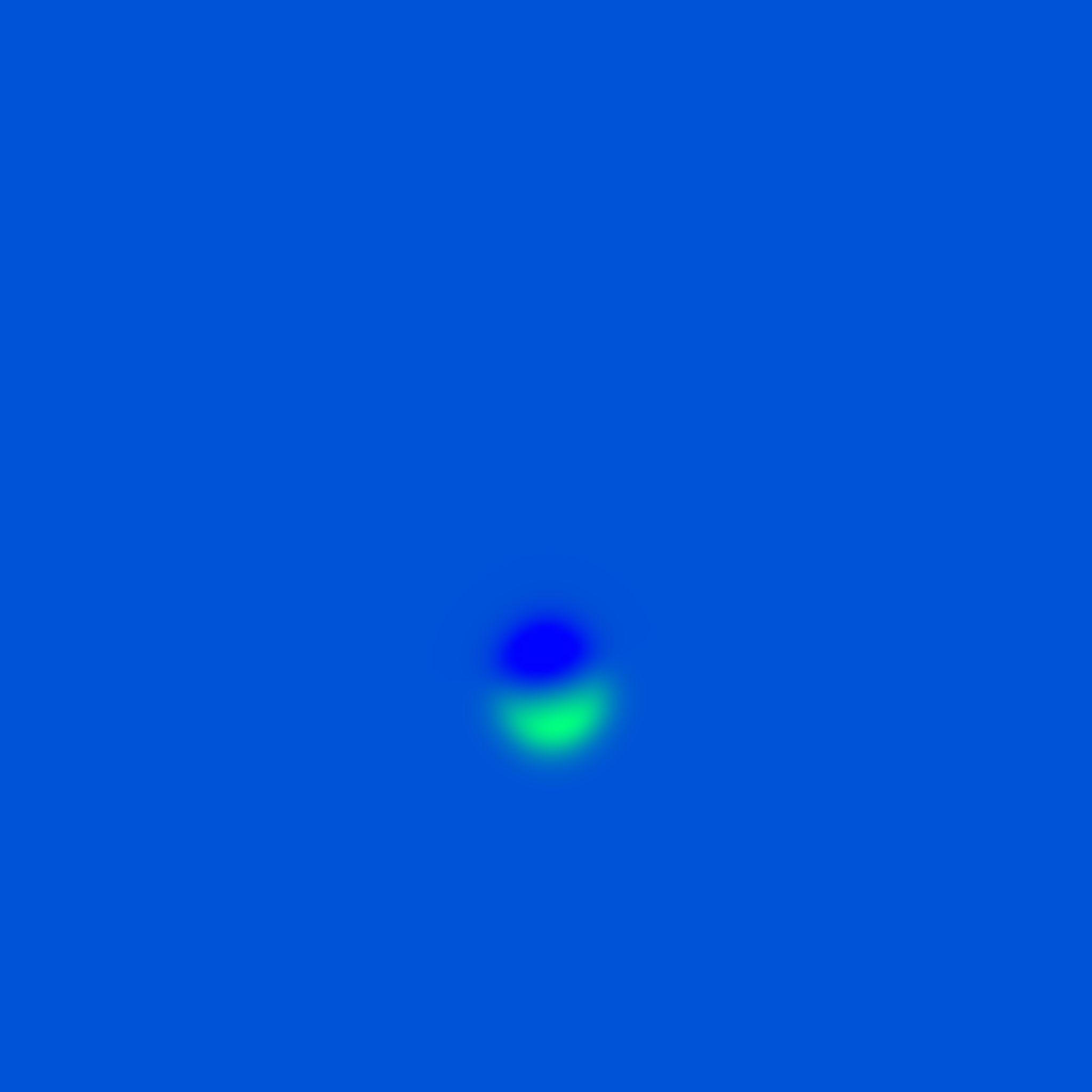}
    \includegraphics[width=0.25\textwidth]{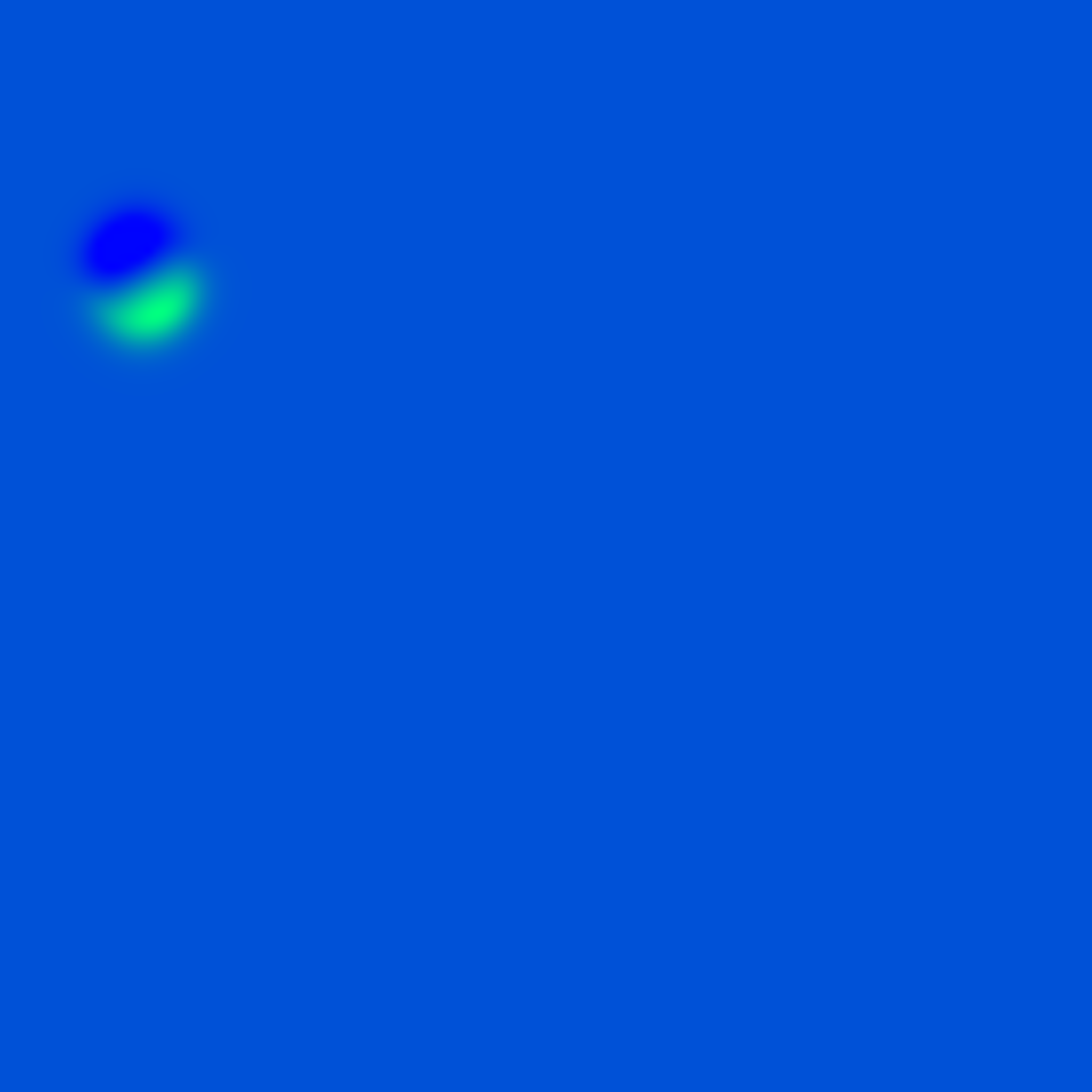}
    \caption{{The top row shows snapshots from two different simulations of the shallow water equations taken at the same time point. These images illustrate the water surface elevation (or height) at that instant, highlighting the markedly different dynamical behaviors that emerge from the two cases. Although both simulations evolve under the same shallow water dynamics, the differences arise solely due to variations in their initial conditions, which are depicted in the bottom row. Note that only the elevation field is shown here; the associated velocity fields are not displayed.}}
    \label{fig:swe}
\end{figure}

Autoencoders have a wide range of applications across various domains due to their ability to learn efficient representations of data. They are commonly used for dimensionality reduction, where high-dimensional data is compressed into a lower-dimensional latent space while preserving essential features. In image processing, autoencoders are employed for denoising, removing noise from corrupted images by learning the underlying clean data distribution \cite{vincent2010stacked}. They are also used in anomaly detection, identifying deviations from patterns in data, such as fraud detection in financial transactions or fault detection in industrial systems \cite{sakurada2014anomaly}. In natural language processing, autoencoders assist in tasks like text generation and sentiment analysis by capturing semantic representations of text \cite{li2015hierarchical}.

\subsection{Variational Autoencoders (VAEs)} \label{subsection:vae}

Variational Autoencoders (VAEs) are a probabilistic extension of standard autoencoders, designed for generative modeling \cite{kingma2022autoencodingvariationalbayes}. The recent surge in generative AI has broadened the scope of VAE applications across diverse domains \cite{Kingma2019}.

Like conventional autoencoders, VAEs define a transformation $\Phi_{V}:\calX \rightarrow \calX$ by mapping inputs into a latent space via a \emph{variational encoder} $e$, and reconstructing them through a \emph{stochastic decoder} $d$. Both $e$ and $d$ are parameterized by deep neural networks with parameters $\theta^e$ and $\theta^d$, respectively. Unlike deterministic autoencoders, the encoder in a VAE produces parameters of a latent distribution. Specifically, $e(z \mid x; \theta^e) \approx p_{\theta^e}(z \mid x)$, where $z \sim \calN(\mu_{\theta^e}(x), \Sigma_{\theta^e}(x))$. Here, the encoder outputs a mean vector and (log-)covariance, from which $z$ is sampled. This distribution, often referred to as the \emph{variational posterior}, allows the model to learn uncertainty-aware latent representations.

The decoder then reconstructs the input from a sample $z$ by generating the parameters of a likelihood distribution $p_{\theta^d}(x \mid z)$, from which a reconstruction $\tilde{x}$ is drawn. In practice, Gaussian or categorical distributions are commonly used depending on the nature of the data. VAE training is based on maximizing the Evidence Lower Bound (ELBO), which approximates the intractable log-likelihood:
\begin{align}
    p_{\theta^e, \theta^d}(x) = \int_{\calZ} p_{\theta^d}(x \mid z) p(z) \, \text{d}z \geq \text{ELBO}(x; \theta^e, \theta^d).
\end{align}
For a given sample $x \sim \calX$, the ELBO takes the form \cite[Sect. 4.1]{GenModIntro}, \cite[Sect. 21.2.4]{murphy2023probabilistic}
\begin{equation}\label{eq:ELBO}
    \operatorname{ELBO}(x; \theta^e, \theta^d) = \mathbb{E}_{z \sim e(z \mid x; \theta^e)}\left[-\log p_{\theta^d}(x \mid z)\right] + \operatorname{KL}\left(e(z \mid x; \theta^e) \, \| \, p(z)\right),
\end{equation}
where $\operatorname{KL}(\cdot \| \cdot)$ denotes the Kullback-Leibler divergence.

To enable backpropagation through stochastic nodes, VAEs employ the \emph{reparameterization trick} \cite{kingma2015variationaldropoutlocalreparameterization}
\begin{equation}
    z = \mu_{\theta^e}(x) + \exp(\Sigma_{\theta^e}(x)) \odot \epsilon, \quad \epsilon \sim \calN(0, \operatorname{I}),
\end{equation}
where $\odot$ is the element-wise multiplication, this formulation ensures differentiability by expressing $z$ as a deterministic function of $x$ and a noise term. Typically, a diagonal covariance structure is assumed, and $\Sigma_{\theta^e}(x)$ outputs log-variances.

When a Gaussian likelihood is chosen:
\begin{equation}
    p_{\theta^d}(x \mid z) = (2\pi \sigma)^{-d/2} \exp\left(-\frac{1}{2\sigma} \|d(z; \theta^d) - x\|_2^2 \right),
\end{equation}
the log-likelihood simplifies to a squared error form
\begin{equation}
    \log p_{\theta^d}(x \mid z) = -\frac{1}{2\sigma} \|d(z; \theta^d) - x\|_2^2 + \text{const}.
\end{equation}

Thus, the ELBO used in our experiments becomes
\begin{equation}\label{eq:ELBO_experiments}
    \operatorname{ELBO}(\theta^e, \theta^d) = \mathbb{E}_{x \sim \calX} \left( \mathbb{E}_{z \sim e(z \mid x; \theta^e)} \left(\frac{1}{2\sigma} \|d(z; \theta^d) - x\|_2^2 \right) + \operatorname{KL}(e(z \mid x; \theta^e) \| p(z)) \right),
\end{equation}
assuming a standard normal prior $p(z) = \calN(0, \operatorname{I})$.

Alternative objectives and model variants have been proposed. Wasserstein Autoencoders (WAEs), for instance, minimize the Wasserstein distance between the empirical and model distributions \cite{tolstikhin2019wassersteinautoencoders}. Latent structures such as sparsity can be induced using spike-and-slab distributions \cite{pmlr-v115-tonolini20a}, and hierarchical or flow-based posterior priors provide richer expressiveness \cite{vahdat2021nvaedeephierarchicalvariational, tomczak2018vaevampprior, chen2017variationallossyautoencoder, tomczak2022deep}.

VAEs are widely used in solving inverse problems. Applications include posterior approximation for uncertainty quantification \cite{goh2019solving}, regularization in inverse solvers \cite{doi:10.1137/21M140225X, Prost2023}, and generative tasks such as image synthesis and molecular design \cite{kingma2013auto}. Their versatility makes them integral to both scientific and industrial workflows.

\section{Paired Autoencoders}\label{sec:pair}

Paired autoencoders present a unique architectural approach for tackling inverse problems by leveraging latent space representations to establish mappings between input and target spaces. This strategy shares conceptual similarities with various methodologies across different domains, including operator learning and reduced order modeling \cite{lee2020model}, both of which utilize latent spaces to construct surrogate operators and reduced models. In the specific context of linear autoencoders and linear mappings, paired autoencoders exhibit a close relationship with principal component analysis (PCA) \cite{benner2015survey,antoulas2020interpolatory}.

The broader landscape of inverse problems has seen extensive adoption of neural networks, addressing diverse tasks ranging from full end-to-end inversion \cite{kulkarni2016reconnet} and regularization \cite{afkham2021learning,li2020nett} to uncertainty quantification \cite{goh2019solving,lan2022scaling} and beyond \cite{bai2020deep,lucas2018using}. Paired autoencoders contribute to this field by employing latent structures to learn mappings between input and target spaces in inverse problems, thus exhibiting parallels to techniques in operator learning and model reduction \cite{kovachki2023neural,lee2020model}.

The paired autoencoder framework offers a novel methodology that integrates the strengths of both data-driven and model-based approaches, positioning itself as a powerful tool for solving inverse problems within scientific computing. Notably, paired autoencoders inherently provide a regularizing effect, obviating the need for explicit regularization parameter selection or computationally intensive algorithms. This characteristic enables the efficient representation of the regularized inverse mapping within a lower-dimensional latent space.

\subsection{Setup}\label{subsection:setup}

{Recent research highlights the effectiveness of \emph{paired autoencoders} for solving inverse problems. To the best of our knowledge, the pairing of two autoencoders was first considered by Boink and Brune in \cite{boink2019learned} and subsequently investigated in \cite{chung2024paired, hart2025paired, feng2022intriguing, feng2024auto, wang2024wavediffusion, feng2024hidden, piening2024paired}. This approach uses latent spaces to compactly represent both data and models as discussed in Section~\ref{subsection:autoencoders}, leading to inherent regularization and improved inverse solution quality. By mapping between latent spaces, paired autoencoders facilitate an inversion process. This capability underpins a novel hybrid strategy combining data-driven and model-based methods. The result is efficient, regularized solutions achieved without costly forward computations, the difficulties of non-convex optimization, or the computational overhead of explicit regularization parameter tuning.}

Mathematically, the encoder $e_x\colon \mathcal{X} \to \mathcal{Z}_x$ maps the quantity of interest $x \in \mathcal{X}$ to a latent representation $z_x \in \mathcal{Z}_x$, while the decoder $d_x\colon \mathcal{Z}_x \to \mathcal{X}$ reconstructs the quantity of interest from its latent space representation $z_x$. Similarly, for the data space, the encoder $e_y\colon \mathcal{Y} \to \mathcal{Z}_y$ maps $y \in \mathcal{Y}$ to $z_y \in \mathcal{Z}_y$, and the decoder $d_y\colon \mathcal{Z}_y \to \mathcal{Y}$ aims to reconstruct $y$. Both autoencoders are parameterized by network parameters $\theta_x = (\theta_x^{\text{e}}, \theta_x^{\text{d}})$ and $\theta_y = (\theta_y^{\text{e}}, \theta_y^{\text{d}})$, respectively. The reconstruction error is measured using a loss function $\mathcal{J}$, such as the mean squared error (MSE) or cross-entropy loss. The objective for each autoencoder is to find network parameters $\theta_x$ and $\theta_y$ that minimize some loss between the input $x$ and $y$ and their corresponding mappings $\tilde x = (d_x \circ e_x)(x)$ $\tilde y = (d_y \circ e_y)(y)$, i.e.,
$$
\mathbb{E}_{x\sim \pi_x} \ \mathcal{J}((d_x \circ e_x)(x),x) \qquad \text{and} \qquad \mathbb{E}_{y\sim \pi_y} \ \mathcal{J}((d_y \circ e_y)(y),y).
$$
Here $\pi_x$ is the probability distribution of the quantity of interest and $\pi_y$ of the data. Note that both objectives are independent optimization problems and therefore can be trained independently. 

However, the key element in paired autoencoders is the integration of connecting latent space mappings. We define a latent space forward operator as $M:\mathcal{Z}_x\to \mathcal{Z}_y$ and a latent space inverse operator $M^\dagger$ as $M^\dagger:\mathcal{Z}_y\to \mathcal{Z}_x$. Note that, since each of the encoders $e_x$ and $e_y$ are a latent space representation of the data and the corresponding QoI, $M$ and $M^\dagger$ can be interpreted as the forward and inverse process within the latent space mimicking $F$ and $F^\dagger$. The composite function $d_y \circ M \circ e_x$ is a surrogate for the forward map $F$ with predicted observation $\hat y = (d_y \circ M \circ e_x)(x)$. Similarly, a surrogate inverse map of $F^\dagger$ can be written as $\approx d_x \circ M^\dagger \circ e_y$. With given observations $y$, a predicted quantity of interest is given as 
$$\hat x = (d_x \circ M^\dagger \circ e_y)(y).$$

The latent space mappings $M$ and $M^\dagger$ are trainable and parameterize by $\theta_M$ and $\theta_{M^\dagger}$ respectively, see Figure~\ref{fig:pair}. For instance, one might choose deep neural networks for intricate mappings or use just linear transformations. Non-trainable mappings are also an option, such as the identity mapping, which effectively makes both autoencoders to share the same latent space, i.e., $\mathcal{Z}_x = \mathcal{Z}_y$. By reducing the trainable parameters through this simplification, we observe that these setups perform effectively in practice, as illustrated in \cite{Chung_2024}. However, making the latent spaces identical may not always be beneficial for performance, as empirically observed in other experiments, see Section~\ref{subsection:imagerestore}. Empirically, it has been shown that linear mappings are sufficient to capture the latent space interactions well. Note if we assume, as above, the linear mapping of the autoencoders, we are able to investigate the optimality of the latent space mappings as shown in \cite{hart2025paired} {and stated in Theorem~\ref{theorem:linmap}}.

Another design choice is how to measure the loss, including the latent space mappings, to train the latent space parameters  $\theta_M$ and $\theta_{M^\dagger}$. For instance, one may directly include mappings
\begin{equation}\label{eq:zx2zy}
    \mathbb{E}_{z_x\sim \pi_{z_x}} \mathcal{J}(M(z_x), z_y) \qquad \text{and/or}\qquad \mathbb{E}_{z_y \sim \pi_{z_y}} \mathcal{J}(z_x, M^\dagger (z_y))
\end{equation}
or full mappings such as
\begin{equation}\label{eq:surrogates}
\mathbb{E}_{x \sim \pi_x} \mathcal{J}(d_y \circ M \circ e_x (x), y) \qquad \text{and/or}\qquad \mathbb{E}_{y\sim \pi_y} \mathcal{J}(d_x \circ M^\dagger \circ e_y (y), x),
\end{equation}
while other loss objectives are possible. Nevertheless, a full objective is a combination of the objective terms, for instance
\begin{align}
    \min_\theta \  &\alpha_x \mathbb{E}_{x\sim \pi_x} \ \mathcal{J}((d_x \circ e_x)(x),x) \label{eq:PAEtrainfull}\\
    & \qquad +  \alpha_y \mathbb{E}_{y\sim \pi_y} \ \mathcal{J}((d_y \circ e_y)(y),y) \nonumber\\
    & \qquad \qquad + \alpha_M \mathbb{E}_{x\sim \pi_x} \mathcal{J}(d_y \circ M \circ e_x (x), y) \nonumber\\
    & \qquad \qquad \qquad + \alpha_{M^\dagger}\mathbb{E}_{y\sim \pi_y} \mathcal{J}(d_x \circ M^\dagger \circ e_y (y), x) \nonumber
\end{align}
with weights $\alpha_x, \alpha_y, \alpha_M, \alpha_{M^\dagger}>0$ and where $\theta = [\theta_x^e, \theta_x^d, \theta_y^e, \theta_y^d, \theta_M, \theta_{M^\dagger}]$ is a collection of all network parameter. 

If one considers to just learn the forward mapping we may not need to learn the inverse mapping $M^\dagger$ and vice versa. Some methods propose a two-stage training process: first training each autoencoder separately, and then learning a mapping between their latent spaces. However, if enough data is available, training the autoencoders and the mapping simultaneously is advantageous. This joint approach ensures the latent spaces are learned in a way inherently suited for the subsequent mapping task (whether inversion or a forward operation), rather than being developed without considering it. Given our objective of constructing surrogate forward and, in particular, inverse mappings, we advocate for the loss function presented in Equation~\ref{eq:surrogates} and Equation~\ref{eq:PAEtrainfull}, since it explicitly targets the desired properties of the surrogate models.

\begin{figure}
    \centering
    \begin{tikzpicture}

	\node[fill=python1!50, minimum width=3.0cm, minimum height=0.5cm] (x) at (0,0) {$x$};
    \draw[fill=python9!50,draw=none]
             ([yshift=-0.2cm,xshift=0.0cm]x.south west) -- ([yshift=-1.5cm,xshift=0.75cm]x.south west) -- ([yshift=-1.5cm,xshift=-0.75cm]x.south east) -- ([yshift=-0.2cm,xshift=0.0cm]x.south east) -- cycle node (e) at ([yshift=-0.75cm,xshift=0.0cm]x.south) {$\begin{matrix}\mbox{encoder} \\[-0.0ex] e_{x} \end{matrix}$}; 
               
    \node[fill=python3!50, minimum width=1.5cm, minimum height=0.5cm] (zx) at (0,-2.3) {$z_x$};    
    \draw[fill=python4!50,draw=none]
             ([yshift=-0.2cm,xshift=0.0cm]zx.south west) -- ([yshift=-1.5cm,xshift=-0.75cm]zx.south west) -- ([yshift=-1.5cm,xshift=0.75cm]zx.south east) -- ([yshift=-0.2cm,xshift=0.0cm]zx.south east) -- cycle node (d) at ([yshift=-0.75cm,xshift=0.0cm]zx.south) {$\begin{matrix}\mbox{decoder} \\[-0.0ex] d_{x} \end{matrix}$};  
    \node[fill=python1!50, minimum width=3.0cm, minimum height=0.5cm] (xt) at (0,-4.5) {$\tilde x$};      

    \node[fill=python2!50, minimum width=2.5cm, minimum height=0.5cm] (b) at (6,0) {$y$};
    \draw[fill=python5!50,draw=none]
             ([yshift=-0.2cm,xshift=0.0cm]b.south west) -- ([yshift=-1.5cm,xshift=0.5cm]b.south west) -- ([yshift=-1.5cm,xshift=-0.5cm]b.south east) -- ([yshift=-0.2cm,xshift=0.0cm]b.south east) -- cycle node (e) at ([yshift=-0.75cm,xshift=0.0cm]b.south) {$\begin{matrix}\mbox{encoder} \\[-0.0ex] e_{y} \end{matrix}$}; 
               
    \node[fill=python6!50, minimum width=1.5cm, minimum height=0.5cm] (zb) at (6,-2.3) {$z_y$};    
    \draw[fill=python9!50,draw=none]
             ([yshift=-0.2cm,xshift=0.0cm]zb.south west) -- ([yshift=-1.5cm,xshift=-0.5cm]zb.south west) -- ([yshift=-1.5cm,xshift=0.5cm]zb.south east) -- ([yshift=-0.2cm,xshift=0.0cm]zb.south east) -- cycle node (d) at ([yshift=-0.75cm,xshift=0.0cm]zb.south) {$\begin{matrix}\mbox{decoder} \\[-0.0ex] d_{y} \end{matrix}$};  
    \node[fill=python2!50, minimum width=2.5cm, minimum height=0.5cm] (bt) at (6,-4.5) {$\tilde  y$}; 

\begin{scope}[-latex,shorten >=9pt,shorten <=9pt,line width=5pt]
    \draw[matlab2!50!matlab3]  ([yshift=-0.2cm]zb.west) to ([yshift=-0.2cm]zx.east);
    \draw[matlab1!50!matlab2] ([yshift=0.2cm]zx.east) to ([yshift=0.2cm]zb.west);    
\end{scope}
\node[matlab1!50!matlab2] (m_forward) at (3.0,-1.7) {$M$};
\node[matlab2!50!matlab3] (m_inverse) at (3.0,-2.9) {$M^{\dagger}$};

\begin{scope}[-latex,shorten >=9pt,shorten <=9pt,line width=5pt]
    \draw[matlab1!50!matlab2] ([yshift=-0.0cm]x.east) to ([yshift=-0.0cm]b.west);
    \draw[matlab2!50!matlab3] ([yshift=0.0cm]bt.west) to ([yshift=0.0cm]xt.east);
\end{scope}

\node[matlab1!50!matlab2] (f_forward) at (3.0,0.3) {$F$};
\node[matlab2!50!matlab3] (f_inverse) at (3.5,-4.8) {$F^{\dagger}$};
     
\end{tikzpicture}
    \caption{Schematic of a paired autoencoder for inverse problems. The $x$-{autoencoder} gives a representation of the QoI, while the $y$-autoencoder provides a representation of the observation space $\mathcal{Y}$. The mappings $M$ and $M^\dagger$ between the latent space $\mathcal{Z}_x$ and $\mathcal{Z}_y$ are surrogate mappings of the forward and the inverse mappings $F$ and $F^\dagger$, respectively. A full surrogate forward mapping and inverse mapping $F \approx (d_y \circ M \circ e_x) (x)$ and $F^\dagger \approx (d_x \circ M^\dagger \circ e_y) (y)$.}
    \label{fig:pair}
\end{figure}
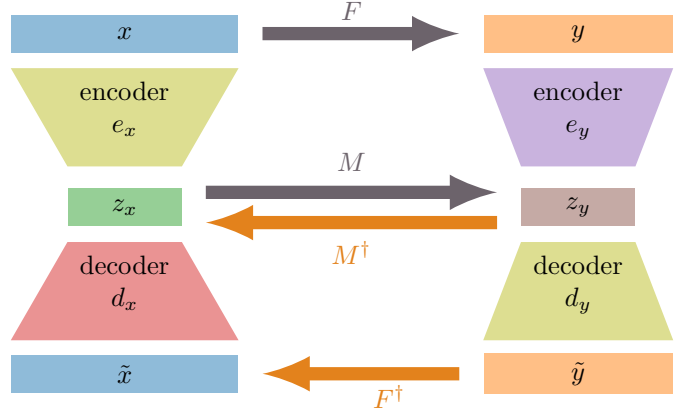

\subsection{Reconstruction Quality} \label{subsection:quality}

{We categorize data into (i) \emph{in-distribution (ID)} data, which is drawn from the same distribution as the training set—typically modeled as $x \sim \pi_x$ with $y = F(x) + \varepsilon$ -- and is expected to be accurately reconstructed if learned properly; (ii) \emph{out-of-distribution (OOD)} data, which originates from a substantially different distribution and typically lies outside the model’s generalization capacity, and therefore \emph{should} exhibit poor reconstruction—accurate reconstruction in such cases may suggest a failure to distinguish distributional shift; a third category is \emph{near-OOD} data, which was not seen during training but lies near the support of the training distribution and may be modeled through mild extrapolation or adaptation \cite{fort2021exploring}. This distinction is crucial for understanding model behavior and robustness, particularly in settings where reconstruction fidelity serves as a proxy for distributional fit.
}

In light of this, data-driven methodologies for inverse problems, such as end-to-end and paired autoencoder architectures, are fundamentally \emph{likelihood-free}. Following the acquisition of the inversion mapping $d_x \circ M^\dagger \circ e_y$ and the determination of its model parameters $\theta$, there is no inherent likelihood function available to evaluate the reconstruction fidelity of novel input data. Only if the forward operator $F$ is available -- which might not be the case -- likelihoods may be accessible, but then the computational cost often prohibits their computation.   Consequently, these methodologies generate deterministic reconstructions without a quantifiable metric for assessing accuracy. This limitation precludes the determination of whether input data originates from the training distribution or is out-of-distribution, thereby rendering the assessment of reconstruction quality indeterminate. In applications where reconstruction accuracy is paramount, this characteristic presents a significant impediment.

Despite its likelihood-free nature, the paired autoencoder framework, however, provides alternative metrics for estimating reconstruction quality through the analysis of its constituent trained mappings and therefore sets itself apart from other data-driven approaches. 

Assuming the paired autoencoders have been successfully trained, yielding the encoders $e_x, e_y$, decoders $d_x, d_y$, along with the forward and inverse latent space mappings $M$ and $M^\dagger$, respectively. One may want to establish accuracy bounds of the forward prediction $\| F(\hat x) - y\|$ and of the predicted quantity of interest $\| \hat x- x \|$. It has been demonstrated {in \cite{boink2019learned, chung2024paired}} that these quantities are bounded by computable expressions. 

These computable expressions may now be indicators that the reconstructed quantity of interest, $\hat x = (d_x \circ M^\dagger \circ e_y) (y)$, may be unreliable, leading us to suspect that an $(x,y)$ sample belongs to OOD data. To be more precise, let us consider one particular bound on $\| \hat x- x \|$.

\begin{thm}[\cite{chung2024paired}]
\label{theorem:error}
    Let $e_y:\mathcal{Y} \to \mathcal{Z}_y$ and $d_x:\mathcal{Z}_x \to \mathcal{X}$ be Lipschitz continuous on their corresponding metric spaces with Lipschitz constants $L_y, L_x \geq0$ each equipped with sub-multiplicative norms and $\|\tilde y- y\| \leq \delta$ for $\delta \geq 0$. Further, assume
    for any $z_y \in \mathcal{Z}_y$ $z_x \in \mathcal{Z}_x$, and $x \in \mathcal{X}$ there exist constants $\xi_y,\xi_M,\xi_x \geq 0$ such that 
    $$\norm{e_y(d_y(z_y)) - z_y} \leq \xi_y, \qquad \| M^\dagger M z_x - z_x \| \leq \xi_M,\qquad\text{and}\qquad\norm{d_x(e_x(x)) - x} \leq \xi_x.$$ 
    Then
    \begin{equation*}
        \norm{\hat x-x} \leq L_x \left(\norm{M^\dagger}  \left(L_y \delta +  \xi_y\right)+\xi_M \right) + \xi_x.
    \end{equation*}
\end{thm}
{The theorem and its proof are provided in \cite{chung2024paired}.}
This bound inherently requires all learned mappings $e_x,e_y, d_x, d_y, M$, and $M^\dagger$. We can see that beyond the Lipschitz constants, the tightness of the bound hinges on the precision of the quantity of interest autoencoder $\xi_x$, the data autoencoder in the latent space $\xi_y$, and how well the latent mappings $\xi_M$ can be inverted. Furthermore, $\delta$ is influenced by the accuracy of the forward surrogate and the noise level in the data.

Notice that $\xi_M$ disappears in the specific scenario where $M^\dagger M = I$. Similarly, if the model and data autoencoders achieve effective compression of their inputs, then $\xi_x$ 
and $\xi_y$ become negligible. Other bounds can be derived, leading to a range of methods for estimating error bounds and thus assessing the quality of the reconstruction. In our numerical experiments, we specifically illustrate the following measures

\begin{align}\label{subsection:metrics}
     \frac{\|(d_y \circ e_y)(y) - y \|_2}{\| y \|_2}, \qquad\qquad      \frac{\|(d_x \circ e_x)(\hat{x}) - \hat{x} \|_2}{\| \hat{x} \|_2}  \\
     \frac{\norm[2]{(d_y \circ M \circ e_x)(\hat x) - y}}{\norm[2]{y}} \\ \quad
    \frac{\norm[2]{(M^{\dagger} \circ e_y)(y) - e_x(\hat{x})}}{\norm[2]{e_x(\hat x)}} \qquad\text{and}\qquad
    \frac{\norm[2]{(M \circ e_x) (\hat x) - e_y(y)}}{\norm[2]{e_y(y)}}.
\end{align}

We can calculate these measures for the training dataset, establishing a baseline distribution. Upon receiving new data, we compute the same measures. If these values fall within the established training distribution, a related theorem suggests that the reconstructed quantity of interest,  
$\hat x$ is likely to be within the expected distribution and accurately recovered. Conversely, if the new data's measures deviate from the training distribution, we anticipate an OOD and potentially poor reconstruction. We exemplify this idea in the image restoration section Section~\ref{subsection:imagerestore}.

Assuming the forward model $F$ is linear, i.e., $F \in \bbR^{m \times n}$ and assuming that each of the autoencoder pieces is also linear according to Section~\ref{subsection:autoencoder}, denoted by $E_x, E_y, D_x$ and $D_y$, we may obtain some theoretical results on the optimality of the latent space mapping.  Here we assume $\varepsilon$ is independent noise with zero mean symmetric positive definite covariance ${\mathbb{E}_{\varepsilon \sim\pi_{\varepsilon}}}(\varepsilon\varepsilon^\top)=\Gamma_\varepsilon=L_\varepsilon L_\varepsilon^\top$.

\begin{thm} \label{theorem:linmap}
Let $x$ be a random variable with finite first moment and symmetric positive definite second moment $\Gamma_x= L_x L_x^\top$. Further, let $y = Fx + \varepsilon$ with linear forward map $F \in \bbR^{m \times n}$ and $\varepsilon$ be independent noise with zero mean symmetric positive definite covariance ${\mathbb{E}_{\varepsilon \sim\pi_{\varepsilon}}}(\varepsilon\varepsilon^\top)=\Gamma_\varepsilon=L_\varepsilon L_\varepsilon^\top$. Given linear autoencoders defined by $E_x, E_y, D_x$ and $D_y$, then if $E_x$ has full row-rank, the optimal linear forward mapping between latent spaces with minimal norm is given by
\begin{equation}\label{eq:linmap}
    \hat M = E_y F \Gamma_x E_x\t \left(E_x \Gamma_x E_x\t\right)^{-1} \in \argmin_M \ {\bbE_{x \sim \pi_x} \norm[2]{M E_x X - E_y F X}^2.}
\end{equation}
If $E_y$ has full row-rank, then the optimal linear inverse map between latent spaces with minimal norm is given by
\begin{equation}\label{eq:linmapInv}
    \hat M^\dagger = E_x \Gamma_x^\top F^\top E_y^\top \left(E_y \Gamma_y E_y^\top \right)^{-1} \in \argmin_{M^\dagger} \ {\mathbb{E}_{x,\varepsilon \sim\pi_x,\pi_{\varepsilon}} \norm[2]{M^\dagger E_y(FX+\varepsilon) - E_xX }^2.}
\end{equation}

\end{thm}
Notice that with these optimal mappings $\hat M$ and $\hat M^\dagger$, surrogate forward and inversion mappings for $F$ and $F^\dagger$ are given by $D_y \hat M E_x$ and $D_x \hat M^\dagger E_y$, respectively. {Note that we did not impose further assumptions on the linear forward operator $F$, which may lead to non-uniqueness of $\hat M$ and $\hat M^\dagger$.} The proof of Theorem~\ref{theorem:linmap} can be found in \cite{hart2025paired}.

\subsection{Latent Space Inference} \label{subsection:lsinf}
Here we discuss inference approaches using the trained paired autoencoders. Therefore, we suppress the network parameter dependency in the notation for readability. After training, the direct and end-to-end style QoI estimation follows as

\begin{equation}\label{eq:PAEdirect}
    \hat x = (d_x \circ M^\dagger \circ e_y)(y),
\end{equation}
which simply involves encoding the data using the data encoder, mapping from the latent data space to the latent model space using $ M^\dagger$, and decoding using the model decoder.

Like end-to-end networks, the paired autoencoders also train to provide a good estimate of $\hat x$ on average, and there is no expectation that any particular $\hat x$ fits the observed data, in the sense that $F(\hat x) \approx y$. This is where a procedure to refine $\hat x$ is required, based on the actual forward operator $F$, or an accurate surrogate model.

Let us define Latent Space Inversion (LSI), a variant of \eqref{eq:ParameterizedInversion} where the inverse problem is parameterized using the model decoder $d_x$ that implicitly carries the model regularization,

\begin{equation} \label{eq:LSI}
    z_{\rm LSI} \in \argmin_{z} \ \thf \norm{(F \circ d_x) (z) - y}^2 + \tfrac{\alpha}{2} \norm{z - z_0}^2.
\end{equation}

{This functional includes neural networks, and we employ automatic differentiation for computation of its gradients. Our numerical experiments use the gradient-based ADAM optimizer \cite{kingma2014adam}.} The optional regularization term $\tfrac{\alpha}{2} \norm{z - z_0}^2$ with $\alpha>0$ controls how much the optimization can change the initial guess

\begin{equation}\label{eq:LSIinitial}
    z_0 \equiv (M^\dagger \circ e_y)(y).
\end{equation}

The final QoI estimate is followed by decoding optimized latent variables,

\begin{equation}\label{eq:LSI_final}
    \hat x_\text{LSI} = d_x(z_\text{LSI}).
\end{equation}

{The LSI \eqref{eq:LSI} thus enhances the data-fitting capabilities of the paired autoencoder framework, although at the computational cost of evaluating the forward operator during inference. This moves the computational cost closer to that of the variational or plug-and-play methods.}

In summary, the paired autoencoders provide a direct estimate $\hat x$ \eqref{eq:PAEdirect}, a decoder to parameterize the inverse problem \eqref{eq:LSI}, and an initial guess for that problem \eqref{eq:LSIinitial}. The initial guess results in empirically better QoI estimates because the starting point is closer to the true model, as compared to starting with a zero or random $z_0$. The initial guess also provides a straightforward way to regularize \eqref{eq:LSI}, where $\alpha$ follows from cross-validation on the training set. Note that having access to just a decoder/generator, as in \eqref{eq:ParameterizedInversion}, does not come with a good initial guess or simple latent-space regularization of the parameterized inverse problem.

\section{Variational forms of Paired Autoencoders}\label{sec:vpae}
Naturally, the question arises whether the paired autoencoders are compatible with the idea of a variational autoencoder and whether it will be a practically feasible approach. Below, we introduce two such approaches: 1) Variational Paired Autoencoders (VPAE) that replace the autoencoder parts of the paired autoencoders with VAEs; 2) a Variational Latent Mapping that replaces the deterministic latent mappings $M$ and $M^\dagger$ with VAEs.

The experiments section provides a first proof-of-concept of the proposed methods.

\subsection{Variational Paired Autoencoders} \label{subsection:vpae}

The construction of the VPAE starts with two encoders that now output two quantities: the mean and log-covariance of the model and data, respectively. Denote the encoders using 
{
\begin{align}
e_x &= (\mu_{x; \theta^{e}_x}(x), \Sigma_{x; \theta^{e}_x}(x)), \\
e_y &= (\mu_{y; \theta^{e}_y}(y), \Sigma_{y; \theta^{e}_y}(y)).
\end{align}
}
Next, we generate (typically) a single sample (implemented via the reparameterization trick) for each according to 
{
\begin{align}
    &z_x \sim \calN(\mu_{x;\theta^{e}_x}(x), \Sigma_{x;\theta^{e}_x}(x)),\\
    &z_y \sim \calN(\mu_{y;\theta^{e}_y}(y), \Sigma_{y;\theta^{e}_y}(y)),
\end{align}
}
and the stochastic decoders $d_x(z_x; \theta^d_x)$ and $d_y(z_y; \theta^d_y)$ generate samples
\begin{align}
    &\tilde x \sim p_{\theta^{d}_x}(x | z_x), \\
    &\tilde y \sim p_{\theta^{d}_y}(y | z_y).
\end{align} 
So far, we have defined just two regular VAEs. To pair them, we introduce latent-space mappings that map mean and covariance estimates from the data side to the model side and vice versa. That is,
{
\begin{align}
    (\hat \mu_y, \hat \Sigma_y) &= M(\mu_{x; \theta^{e}_x}(x), \Sigma_{x; \theta^{e}_x}(x)), \\
    (\hat \mu_x, \hat \Sigma_x) &= M^\dagger(\mu_{y; \theta^{e}_y}(y), \Sigma_{y; \theta^{e}_y}(y)).
\end{align}
}
Then, these estimates generate the samples
\begin{align}
    &\hat z_{x} \sim \calN(\hat \mu_x, \hat \Sigma_x),\\
    &\hat z_y \sim \calN(\hat \mu_{y}, \hat \Sigma_{y}),
\end{align}
Finally, the decoders generate 
\begin{align}
    &\hat x \sim p_{\theta^{d}_x}(x | \hat z_x), \\
    &\hat y \sim p_{\theta^{d}_y}(y | \hat z_y),
\end{align} 
which are the estimated model from the input data and the estimated data from the input model.

The outlined construction of a VPAE leads to four reconstruction losses and four KL-divergence losses. To see this, consider the standard ELBO loss Equation~\ref{eq:ELBO_experiments}. The VPAE uses one ELBO loss for the data reconstruction, one for the model reconstruction, one for the model estimation from data, and one for the data estimation from the model,

\begin{align}\label{eq:VPAE_training}
\mathbb{E}_{ \{x \sim \calX,  y \sim \calY\} } \bigg[& 
    \mathbb{E}_{z_x \sim e_x(z_x | x; \theta^{e}_x)} \Bigg[\frac{1}{2 \sigma} \|d_x(z_x; \theta^{d}_x) - x \|_2^2 \Bigg] + \operatorname{KL}(e_x(z_x | x; \theta^{e}_x) || p(z_x))
    \quad \text{ELBO - data}\\
    +&\mathbb{E}_{z_y \sim e_y(z_y | y; \theta^{e}_y)} \Bigg[\frac{1}{2 \sigma} \|d_y(z_y; \theta^{d}_y) - y \|_2^2 \Bigg] + \operatorname{KL}(e_y(z_y | y; \theta^{e}_y) || p(z_y))
    \quad \text{ELBO - model} \nonumber\\
    +&\mathbb{E}_{\hat z_y \sim M \circ e_x(z_x | x; \theta^{e}_x)} \Bigg[\frac{1}{2 \sigma} \|d_y(\hat z_y; \theta^{d}_y) - y \|_2^2 \Bigg] + \operatorname{KL}(M \circ e_x(z_x | x; \theta^{e}_x) || p(\hat z_y))
    \quad \text{ELBO - model $\rightarrow$ data} \nonumber\\
    +&\mathbb{E}_{\hat z_x \sim M^\dagger \circ e_y(z_y | y; \theta^{e}_y)} \Bigg[\frac{1}{2 \sigma} \|d_x(\hat z_x; \theta^{d}_x) - x \|_2^2 \Bigg] + \operatorname{KL}(M^\dagger \circ e_y(z_y | y; \theta^{e}_y) || p(\hat z_x)) \bigg].
    \quad \text{ELBO - data $\rightarrow$ model}  \nonumber
\end{align}

A simplified illustration is given in Figure~\ref{fig:VPAE_fig}. While minimizing the above expectations during training, we sample pairs of data and models $\{x \sim \calX,  y \sim \calY\}$ in a stochastic approximation fashion. We also use stochastic approximation using a single sample for the sampling of $z_x, z_y, \hat z_x, \hat z_y$. {We recognize that extending the OOD analysis to the variational paired autoencoder is a promising direction that enables a more rigorous treatment of uncertainty quantification; however, we defer this more complex probabilistic investigation to future work.}

\begin{figure}
    \centering
    \input{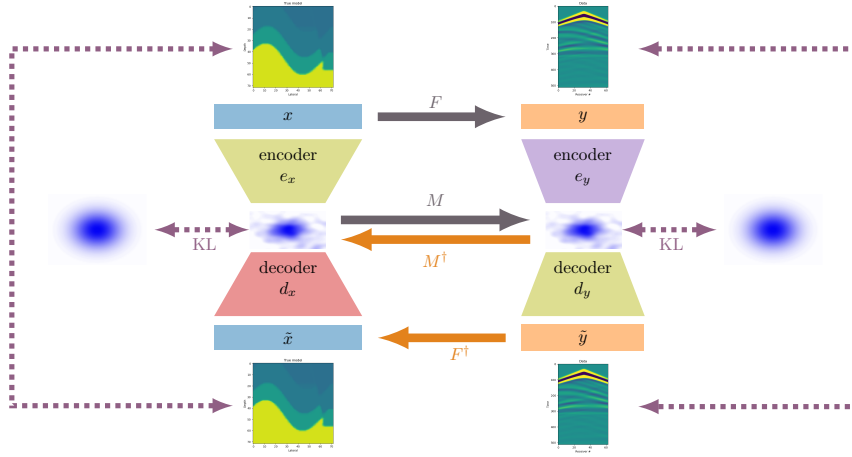}
    \caption{Diagram of the Variational Paired Autoencoder from Equation~\ref{eq:VPAE_training}. The dashed lines represent components contributing to the loss function. The KL divergence term aligns the latent representation distribution with a standard normal distribution.}
    \label{fig:VPAE_fig}
\end{figure}

\subsection{Inference}\label{subsection:vpaeinf}
Several type of model (and data) estimates are available once training of the encoders and decoders via Equation~\ref{eq:VPAE_training} is complete. A direct estimate follows from encoding the data $y$, applying $M^\dagger$ to map the data latents $z_y$ to model latents $z_x$, and finally decoding the result. In summary, the direct QoI estimate adheres to the same expression as in for regular paired autoencoders Equation~\ref{eq:PAEdirect},

\begin{equation}\label{VPAEdirect}
    \hat x = (d_x \circ M^\dagger \circ e_y)(y).
\end{equation}

More interesting is the availability of samples from the learned posterior and decoding those samples. Specifically, consider a data sample $y$ and its encoding $e(y; \theta^e_y) = (\mu_{y; \theta^{e}_y}(y), \Sigma_{y; \theta^{e}_y}(y))$. The application of $M^\dagger$ provides the mean and log-covariance of the latent model distribution as $(\hat \mu_x, \hat \Sigma_x) = M^\dagger(\mu_{y; \theta^{e}_y}(y), \Sigma_{y; \theta^{e}_y}(y))$. Then, we sample the latent space via $\hat z_{x} \sim \calN(\hat \mu_x, \hat \Sigma_x)$, after which decoding provides the reconstructions $\hat x \sim p_{\theta^{d}_x}(x | \hat z_x)$.

So after encoding some data only once, mapping the latents from data to model once, the sampling and decoding repeat, i.e.,
\begin{equation}\label{VPAEsamples}
d_x(\hat z_x; \theta^d_x), \quad \hat z_x \sim M^\dagger \circ e_y(z_y | y; \theta^{e}_y).
\end{equation}
Sufficiently many samples enable basic statistical analyses in terms of, e.g., the mean and standard deviation per pixel. The VPAE enables latent space inversion, just like the regular paired autoencoders in Equation~\ref{eq:LSI}. However, the VPAE offers more choices for the latent-space initial guess. First, the direct estimate (Equation~\ref{VPAEdirect}) from the VPAE can serve as the initial guess. Second, each sample from \ref{VPAEsamples} may serve as the initial guess. Third, we could use the sample mean as the initial guess.

\subsection{Variational Latent Mapping}\label{subsection:vpaelatent}

  In this section, we present an alternative method for combining paired autoencoders with the ideas from VAEs, specifically for regularizing, and sampling from the latent space. Consider introducing randomness into the learnable latent space mappings $M(z_x; \theta_M)$ and $M^{\dagger}(z_y; \theta_{M^\dagger})$ by modeling them as non-automorphic VAEs, while the model and data autoencoders remain deterministic.   

A probabilistic latent mapping yields several options for training. In particular, this mapping can be trained jointly with the data encoders and decoders, where an objective similar to Equation~\ref{eq:ELBO_experiments} could be included as an objective in addition to Equation~\ref{eq:PAEtrainfull}. Alternatively, we can first train both autoencoders independently or as paired autoencoders, followed by training our probabilistic mapping between the latent spaces. This approach was found empirically to give better results and is implemented in Section~\ref{subsection:imagerestore}.   

In either of the training options described above, the latent mappings $M^{\dagger}: \calZ_y \rightarrow \calZ_x$ and $M: \calZ_x \rightarrow \calZ_y$ are represented by the VAE-like encoder and decoder pairs 

\begin{align}
    e_{z_x}(z_x) = (\mu_{\theta^e_{z_x}}(z_x), \Sigma_{\theta^e_{z_x}}(z_x)) \\
    e_{z_y}(z_y) = (\mu_{\theta^e_{z_y}}(z_y), \Sigma_{\theta^e_{z_y}}(z_y)), 
\end{align} 

and $d_{z_x}(z_{z_y})$ and $d_{z_y}(z_{z_x})$. Here, $z_{z_y}$ are the latent variables of latent variables. $M^\dagger(z_y)$ is computed by first sampling $z_{z_y} \sim \calN(z_y ; \mu_{\theta^e_{z_y}}(z_y), \Sigma_{\theta^e_{z_y}}(z_y))$ and then applying the decoder $z_x = d_{z_x}(z_{z_y})$. For simplicity and consistency, we abuse notation and write this process as $M^{\dagger}(z_y)$ despite the sampling procedure that occurs. Consistent with Section~\ref{subsection:autoencoders}, training proceeds by maximizing a version of the ELBO, adapted to the current non-automorphic setting.

Since data compression already occurs during the data encoding process and a high-fidelity mapping is desirable, we opt to use a non-compressive encoder, but this is not required.

The introduction of a latent mapping based on a sampling procedure enables the generation of multiple reconstructions of the same image, which can give insight into performance of the model. Although this method differs from posterior sampling techniques commonly used in Bayesian uncertainty quantification, it can still provide valuable insight into regions where the model exhibits low confidence.  It is apparent from the options discussed in this and in the previous section that there are many potential variants of (variational) paired autoencoders that can be customized and optimized for varied use cases. 

\section{Experiments}\label{section:applications}

\subsection{Image Restoration}\label{subsection:imagerestore}

Image data is often corrupted in inconsistent or unpredictable locations, motivating techniques to \emph{inpaint}/\emph{restore} these inaccuracies using surrounding image and/or thematic information. Image inpainting, also referred to as image interpolation or completion, aims to reconstruct missing or damaged parts of an image in a visually plausible way \cite{elharrouss2020image,bertalmio2000image, zhou1988image}. We focus on image \emph{restoration} or \emph{blind} inpainting, where the locations of corrupted data in the image are considered unknown beforehand.

Image inpainting techniques find wide-ranging applications across various fields. These include restoring old photographs and damaged artwork \cite{banham1997digital}, removing unwanted objects, watermarks, or text from images, and even advanced image editing for creative purposes \cite{chen2022simple, guillemot2013image}. In medical imaging, inpainting can help fill in missing data in scans, while in satellite imagery, it can be used to reconstruct areas obscured by clouds \cite{armanious2019adversarial,shen2008map}.

Significant progress in digital inpainting, which began in the mid-1990s, was achieved through approaches employing total variation and compressed sensing methods \cite{donoho2006compressed, getreuer2012total, chung2022variable}. We demonstrate the utility of paired autoencoders in this application and how this framework may help to provide insight as to whether a sample is likely to have been generated by the same distribution that it was trained on.

In this simulation study, our goal is to reconstruct a clean MNIST image from corrupted versions. We operate under the assumption that the corrupted pixel locations are unknown, so no mask is available. Starting with the clean MNIST $28\times28$-images $(x_i)_{i = 1}^{60,000}$, we constructed a corrupted dataset $(y_i)_{i = 1}^{60,000}$ by applying a Bernoulli process at the pixel level. For each pixel in each image, we set its value to zero with a probability of $p=0.5$. The resulting training data consisted of pairs of clean and corrupted images $(x_i, y_i)_{i = 1}^{60,000}$. We then proceeded to establish two autoencoder models, one trained on the clean data distribution, given by $(d_x \circ e_x)(\mdot)$, and the other on the corrupted data distribution, given by $(d_y \circ e_y)(\mdot)$. For each of these neural networks, we establish a {six-layer CNN encoder} with a latent space dimension of $128$.  {The decoder mirrors this structure with additional layers for upsampling} (see Appendix~\ref{appendix:imagerestoration}).  Linear latent space mapping $M$ and forward and inverse mappings $M^{\dagger}$ are used. The paired autoencoder networks are trained simultaneously. Reconstruction results for ID test data not seen by the network are shown in the first row of Figure~\ref{fig:results_bip}, which demonstrate the model's strong reconstruction performance on ID examples. 

While straightforward end-to-end neural networks $\Phi\colon \mathcal{Y} \to \mathcal{X}$ yield comparable reconstruction quality (results not shown), they lack mechanisms for assessing in-distribution vs.~out-of-distribution data. Here, we illustrate the tools that paired autoencoders provide for this task, as in Section~\ref{subsection:quality}.

To investigate the effect of OOD data, we generated a second test dataset different from the training set. To be more precise, we eliminated random image blocks of pixels and set them to zero, illustrated in Figure~\ref{fig:results_bip}, (second row, middle). The figure also shows (second row, right) that the reconstructions are adequate but fall short of the quality achieved with the ID data.

\begin{figure}
    \centering
    \includegraphics[width=.95\textwidth]{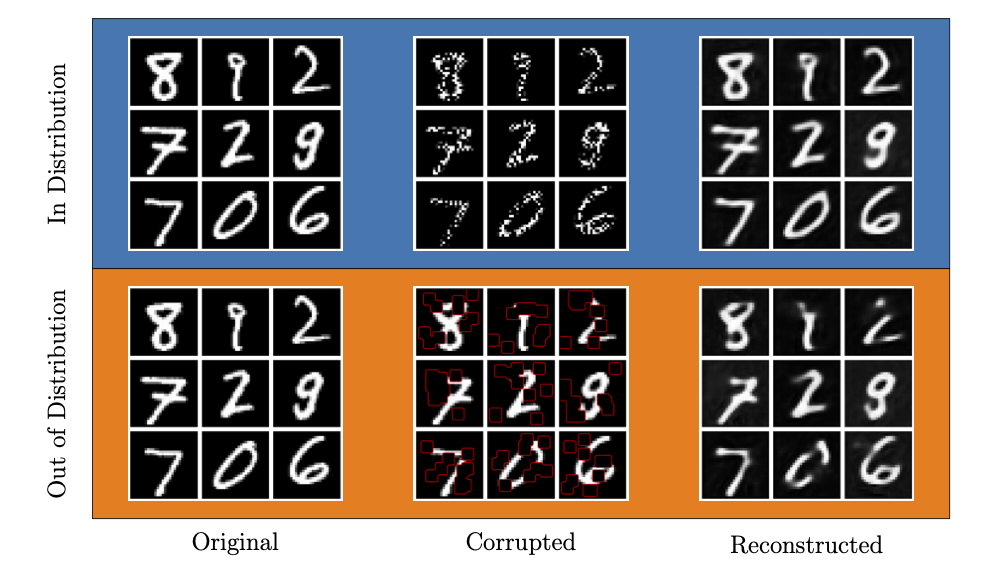}
    \caption{Original data (left), corrupted data (middle), and paired-autoencoder reconstruction (right) for a sample of MNIST images.  Here, in-distribution (ID) samples consist of images with pixels removed randomly, whereas out-of-distribution (OOD) examples have entire chunks of the image removed randomly. Chunks of pixels removed are outlined in red.}
    \label{fig:results_bip}
\end{figure}

As detailed in \cite{chung2024paired} and further discussed in Section~\ref{subsection:quality}, we utilize the metrics outlined in Section~\ref{subsection:metrics} to evaluate the likelihood of an image belonging to the training data distribution. Figure~\ref{fig:hist_bip} displays histograms of these metrics for the ID and OOD samples. While some histograms reveal a distinct difference between the distributions, others show more subtle variations. However, these histograms represent only a one-dimensional projection and thus are useful as a tool to distinguish distributions as opposed to evaluating model performance. A 2D scatter plot further illustrates this distinction (Figure~\ref{fig:scatter_bip}). Notably, plotting the first metric from Figure~\ref{fig:hist_bip} against the second metric clearly reveals a good separation of the clusters. In conclusion, the model's performance is suboptimal for OOD examples, but the metrics offer a valuable indicator of the expected reconstruction quality for new, unseen data.

\begin{figure}
    \centering
    \includegraphics[width=0.95\linewidth]{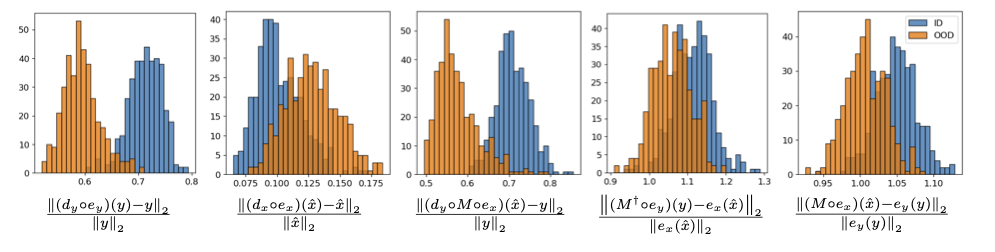}
    \caption{Histogram of paired-autoencoder reconstruction metrics for ID and OOD data computed across 320. For many of the metrics, there is a clear distinction between distribution across the ID and OOD samples.}
    \label{fig:hist_bip}

\end{figure}

\begin{figure}
    \centering
    \includegraphics[width=0.7\linewidth]{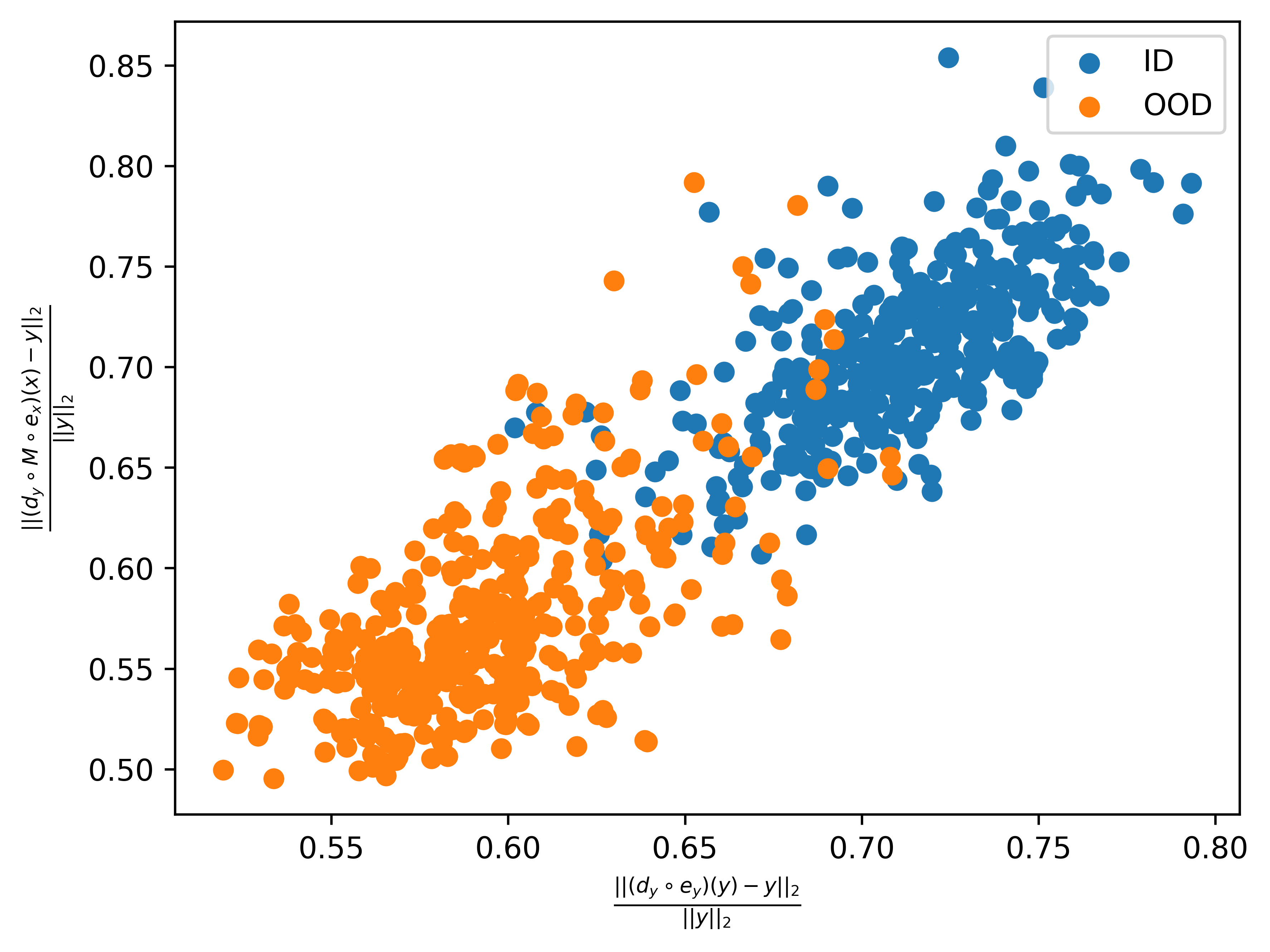}
    \caption{First and third metric of Figure~\ref{fig:hist_bip} plotted against each other to show clustering inter-distribution clustering of data.}
    \label{fig:scatter_bip}
\end{figure}

\subsection{Image Restoration: Variational Latent Mapping}

To gain insight into uncertainties of the reconstructions, we can augment the experiment performed in Section~\ref{subsection:imagerestore} by introducing a probabilistic mapping between the latent spaces as in Section~\ref{subsection:vpaelatent}. 

For the sake of demonstration, a variation of the experiment described in Section~\ref{subsection:imagerestore} was conducted using the MNISTfashion dataset.  Starting with the default $28 \times 28$ images in the MNISTfashion dataset, $(x_i)_{i = 1}^{30,000}$, we generated a corrupted dataset $(y_i)_{i = 1}^{30,000}$ by randomly deleting five $8\times 8$ blocks of pixels, consistent with the generation of the OOD samples in Section~\ref{subsection:imagerestore}. 

To reconstruct an image from a corrupted image, the mapping $M^\dagger$ from $z_y$ to $z_x$ must be learned.  After training paired autoencoders for both datasets with latent dimension $128$, the resulting encoders were used to generate a training set for the variational latent mapping for a new set of images not used in training, $(z_y^{(i)}, z_x^{(i)})_{i = 1}^{30,000}$. 
The latent map consisted of a non-compressive linear variational encoder and linear deterministic decoder, implemented as 2-layer linear neural networks with a hidden dimension of 128.  

Examples of reconstructions generated using this approach are shown in Figure~\ref{fig:VAELatentMap}. We can analyze the model's decision-making process by examining the rightmost column. Notably, the pixels with the highest variance tend to appear along the edges of the image and in areas where the original image was either inherently dark or had sections removed. For example, in the third image, we observe high uncertainty values in the shoe, particularly in regions where fragments were deleted, as well as in naturally black areas of the image. This behavior aligns with expectations, given that the model was trained to restore missing regions in an unsupervised fashion. The elevated uncertainty in these areas reflects the learned sensitivity of the model to occlusions and ambiguous content during reconstruction.

Furthermore, the mean and variance plots illustrate that the model has learned to ignore fine details in the interior regions of clothing items. This is most evident in the fourth image, where the intricate features of the shirt are absent in the mean reconstruction. The corresponding variance plots support this interpretation, showing relatively low uncertainty in those regions, indicating that the model removed these details with high confidence.  This gives insight into the training of the model and suggests that either the training or model hyperparameters need to be adjusted to capture these features of the image.

\begin{figure}
    \centering
    \includegraphics[width=\linewidth]{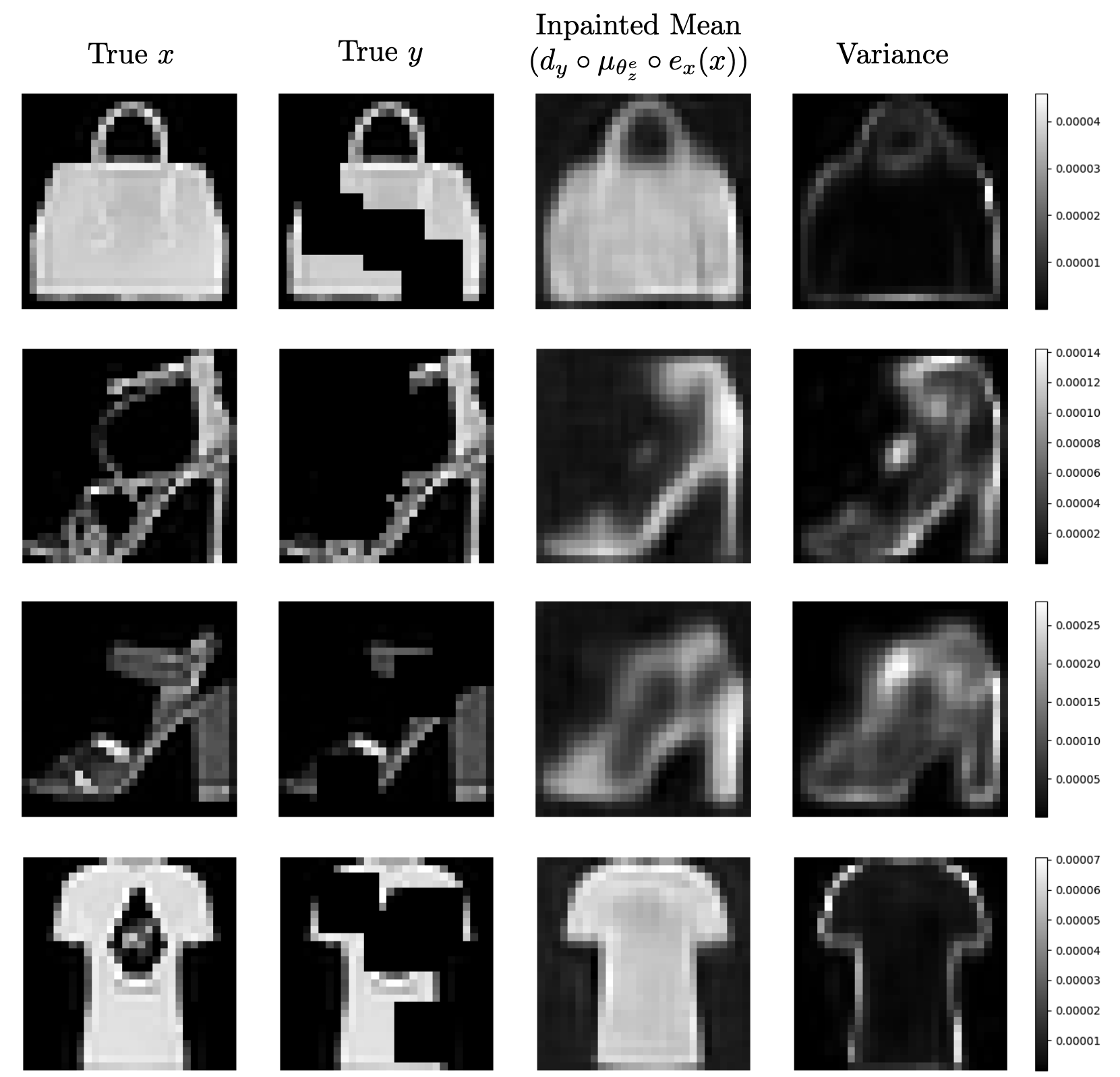}
\caption{Visualization of paired autoencoders with variational mapping between latent spaces.  The leftmost column visualizes the original image $x$ followed by the corrupted image $y$.  The next panel shows the mean of the latent distribution computed by the variational encoder, then passed through the decoder.  Finally, the rightmost column shows the pixel-wise variance of the reconstructions computed across $100$ decoded samples from $M^\dagger (z_y)$ {paired with a colorbar indicating the scale of the variance.  All other images are normalized.}  }
\label{fig:VAELatentMap}
\end{figure}

\subsection{Seismic data inversion}\label{subsection:seismic}
Seismic data provide excellent depth-penetration and high-resolution imaging results to study the Earth's interior, and with this, bears many similarities to medical tomography \cite{haber2014computational}. Active-source seismic data are collected using a collection of seismic sources, $s(\chi,t)$, located at spatial locations $\chi$, and emit a time $(t)$ dependent signature. Multiple receivers measure the pressure (or particle velocity) of the spatio-temporal wave field $u(\chi,t)$ at locations encoded in the selection matrix $P$ as in the introduction, see Figure~\ref{fig:FWI_data_model} for an example. The goal is to estimate the acoustic velocity $x$ given noisy measurements $(P \circ u)(\chi,t) +\varepsilon$. 

The acoustic wave equation,
\begin{equation}\label{acousticWE}
    \nabla^2 u(\chi,t)  - \frac{1}{x(\chi)} \partial_{tt} u(\chi,t) =  s(\chi,t),
\end{equation}
relates the above quantities. Seismic Full-Waveform Inversion (FWI) is the most common approach to formulating the inverse problem. FWI is equivalent to eliminating the PDE constraints and minimizing the unconstrained problem Equation~\ref{eq:VariationalInverse}. 

\begin{figure}
    \centering

     \includegraphics[width=0.35\textwidth]{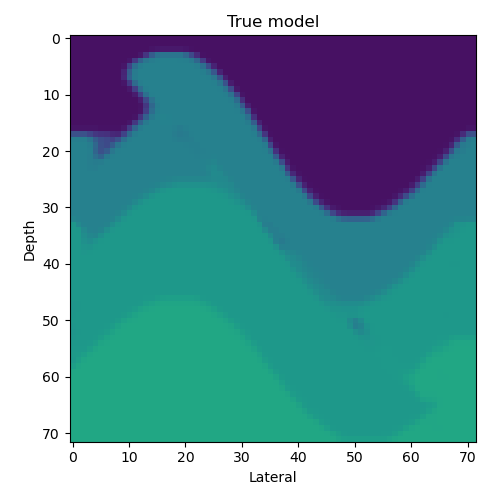}\\
    \includegraphics[width=0.31\textwidth]{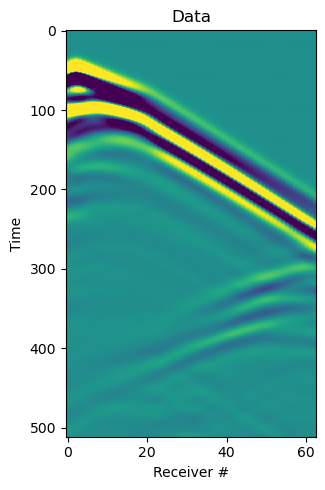}
    \includegraphics[width=0.31\textwidth]{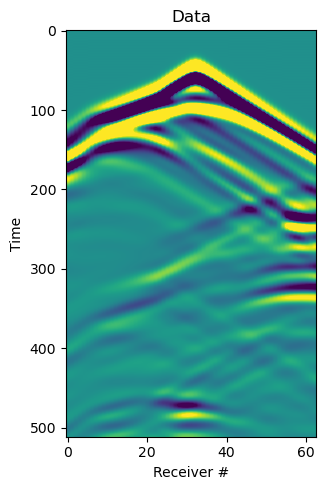}
    \includegraphics[width=0.31\textwidth]{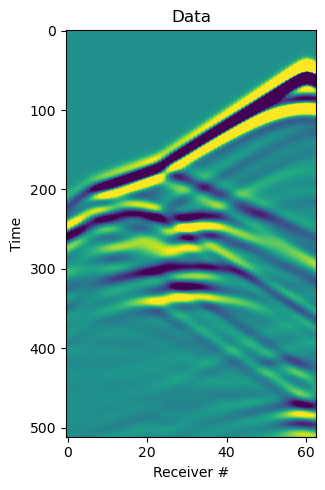}

    \caption{(top) an acoustic velocity model. (bottom) observed time-space data for three out of the $30$ source positions (noise free).} \label{fig:FWI_data_model}
\end{figure}

In this experiment, the observed seismic wavefield data consist of $63$ receivers, $30$ sources, and $512$ time samples for each of the 48,000$/$6,000 training/validation examples of size $700\rm{m} \times 700 \rm{m}$ from the OpenFWI CurveFaultA dataset \cite{deng2022openfwi}. All sources and receivers are located near the top of the computational domain. We used DeepWave \cite{richardson_alan_2023} to generate seismic data. 

It is well known in seismic FWI that having the lowest frequencies available in the observed data ($\lessapprox 3$Hz) enables physics-based inversion to reconstruct nearly all structures, and the availability of the higher frequencies provides the high-resolution details. The importance of the lowest frequencies for physics-based inversion is highlighted by quests to design physical sources specifically for low frequencies \cite{Wolfspar} and techniques that extrapolate seismic data to lower frequencies \cite{LowFreqExtrapolation1,LowFreqExtrapolation2,LowFreqExtrapolation3}.

Therefore, researchers almost always add at least some random (Gaussian) noise to the synthetic data before solving the inverse problem. However, the magnitude of the noise varies wildly in deep-learning studies for seismic FWI. The authors in \cite{wiser} use a Signal-to-Noise-Ratio (SNR) of $12$dB and $0$dB for different experiments, while \cite{zhu2022integrating}  report SNRs of $0$dB and $6$dB. \cite{gupta2024unified} use a (Peak-Signal-to-Noise-Ratio) PSNR ranging approximately between $ 39$ and $84$dB. Yet, other studies do not report any noise \cite{Chung_2024,wang2024wavediffusion}. Here, we take a somewhat middle ground and add zero-mean Gaussian noise (SNR of $30$dB / PSNR $\approx35$dB) to the synthetic data for training and we validate using various noise levels between $0$dB and $70$dB. Figure~\ref{fig:SeismicSpectra} shows some samples of the data and its frequency spectra to illustrate that the noise strongly affects the lowest $\lessapprox 3$Hz and highest frequencies.

\begin{figure}
    \centering
    \includegraphics[width=0.9\linewidth]{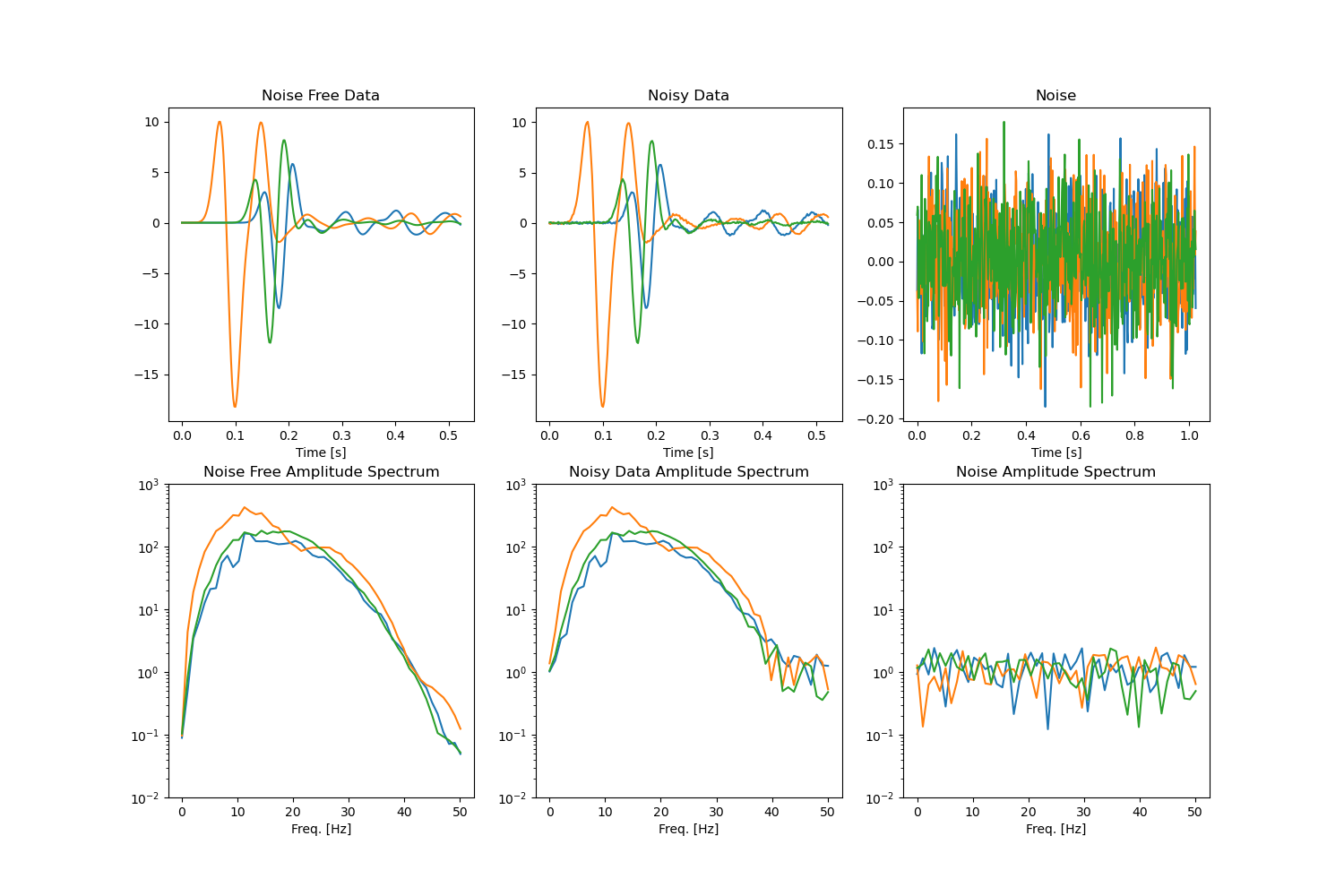}
    \caption{Three time series and their corresponding amplitude spectra for the noise level in the training set(SNR=$30$dB). Validation experiments use various noise levels, see Figures \ref{fig:FWInoisetest} and \ref{fig:SNR10validationex}.}
    \label{fig:SeismicSpectra}
\end{figure}

The paired autoencoders provides an approach similar to Equation~\ref{eq:ParameterizedInversion} by using a model decoder $d_x(z_x,\theta_x^{\rm{ d}})$ to parameterize the inverse problem. What sets the paired autoencoder apart from obtaining a parameterization from a regular VAE decoder or other generative models, is the availability of an accurate initial guess in the latent space, \eqref{eq:LSIinitial}, which we use to both warm start and regularize the refinement problem from Equation~\ref{eq:LSI}. See Appendix ~\ref{appendix} for the network designs used. Numerical experiments show that the decoded refinement result \eqref{eq:LSI_final} reduces the data misfit $\| F(\hat x) - y \|$ and the QoI error in terms of structural similarity index (SSIM) and relative error (RE) $||\hat x - x_\text{true} ||_2 / ||x_\text{true} ||_2$. Figure~\ref{fig:FWIresults}  and Table~\ref{tab:PAEFWIstats} show the merit of latent-space inversion with an accurate initial guess for a single validation model.

\begin{figure}
    \centering
    \scalebox{0.7}{
    \begin{tabular}{cccc}
         \multicolumn{4}{c}{
    \includegraphics[width=0.35\linewidth]{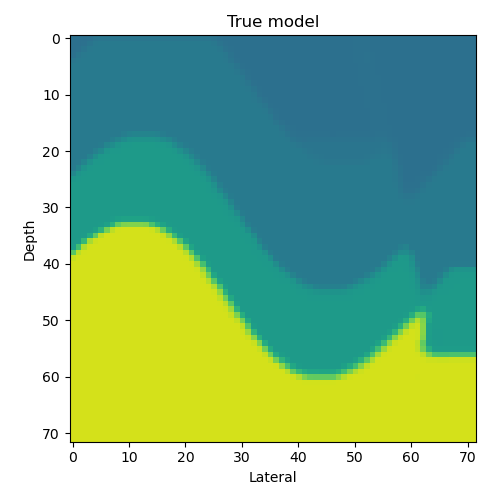}  \includegraphics[width=0.41\linewidth]{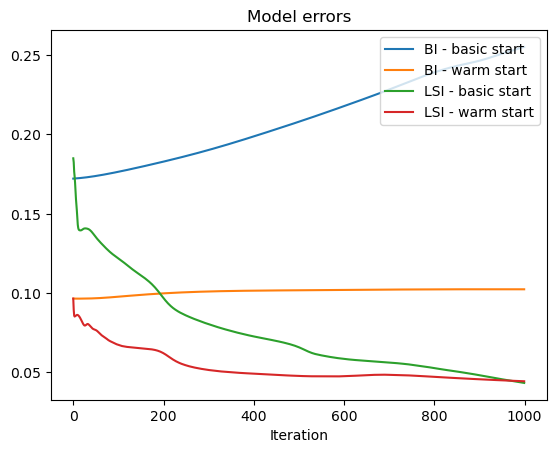}} \\
    basic inversion \eqref{eq:VariationalInverse} & basic inversion \eqref{eq:VariationalInverse} & latent-space inversion \eqref{eq:LSI} & latent-space inversion \eqref{eq:LSI}\\
    basic start  & warm start \eqref{eq:PAEdirect} & basic start & warm start \eqref{eq:LSIinitial}\\
     \includegraphics[width=0.22\linewidth]{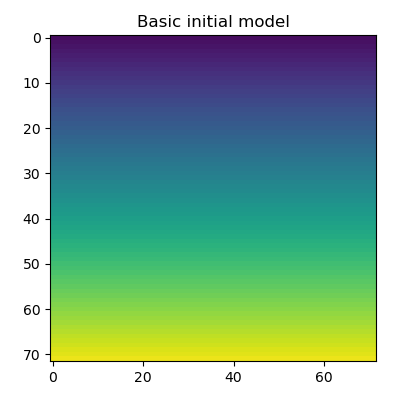}    &      \includegraphics[width=0.22\linewidth]{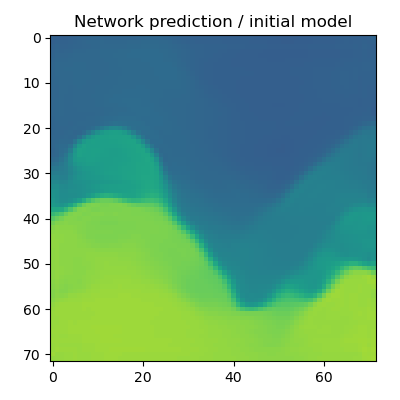} &      \includegraphics[width=0.22\linewidth]{ex3basic_starting} & \includegraphics[width=0.22\linewidth]{ex3network_prediction} \\
      \includegraphics[width=0.22\linewidth]{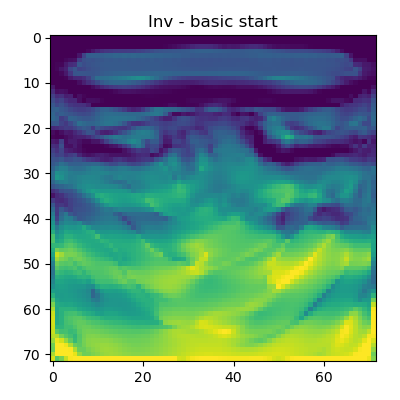}   &       \includegraphics[width=0.22\linewidth]{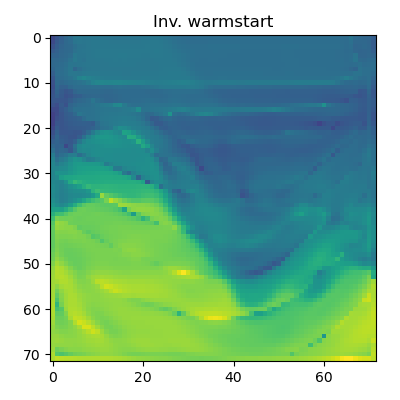} &       \includegraphics[width=0.22\linewidth]{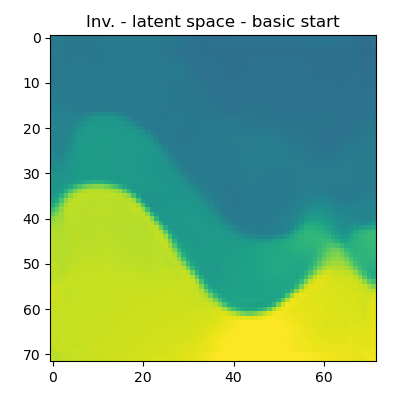} &       \includegraphics[width=0.22\linewidth]{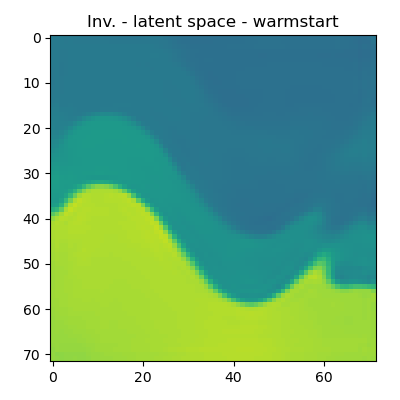} \\
      rel. error = 0.26 & rel. error= 0.12 & rel. error = 0.06 & rel. error = 0.05\\
      SSIM = 0.22 & SSIM = 0.53 & SSIM = 0.92 & SSIM = 0.95\\
       \includegraphics[width=0.25\linewidth]{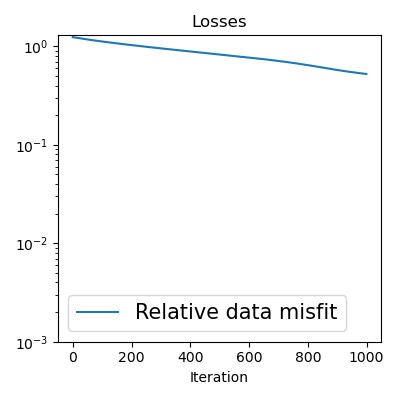}  &        \includegraphics[width=0.25\linewidth]{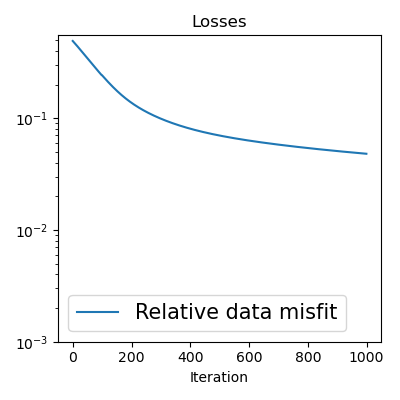} &        \includegraphics[width=0.25\linewidth]{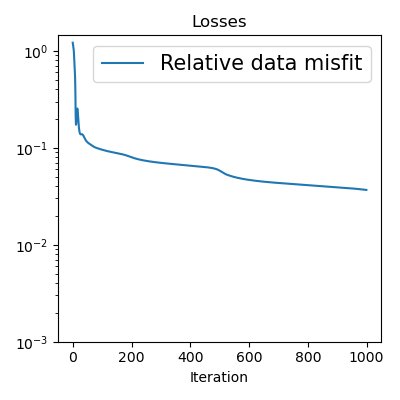} &        \includegraphics[width=0.25\linewidth]{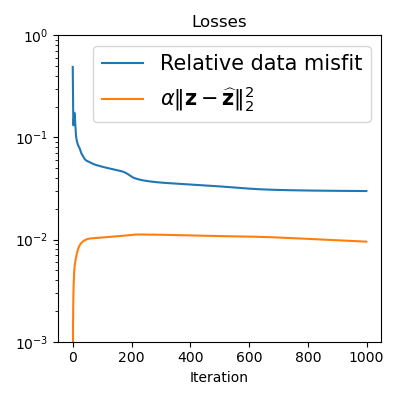}\\
    \end{tabular}
    }
    \caption{All four compared seismic FWI methods reduce the data misfit somewhat, but the result in model estimation varies enormously. The results also show that a direct end-to-end style estimate may not be sufficient when significant noise is present, and the latent-space inversion is required. The availability of the latent-space initial guess both speeds up the inversion and improves the final model accuracy.}
    \label{fig:FWIresults}
\end{figure}

To further investigate the effect of noise on the performance of the direct estimate from the paired autoencoders \eqref{eq:PAEdirect}, and the subsequent LSI \eqref{eq:LSI}, we repeat the above experiment for various noise levels. The autoencoders are the same as above: trained using data with an SNR of $30$. Figure~\ref{fig:FWInoisetest} shows that direct prediction is somewhat impacted by higher noise levels, but subsequent LSI shows to be largely immune to the noise levels selected for this example. { Figure \ref{fig:SNR10validationex} shows another two validation examples using data that contain more noise compared to training data.}

\begin{figure}
    \centering
    \scalebox{0.65}{
    \begin{tabular}{cccc}
    warm start \eqref{eq:LSIinitial} - SNR=0  & warm start \eqref{eq:LSIinitial} - SNR=10 & warm start \eqref{eq:LSIinitial} - SNR=30 & warm start \eqref{eq:LSIinitial} - SNR=70\\
     \includegraphics[width=0.23\linewidth]{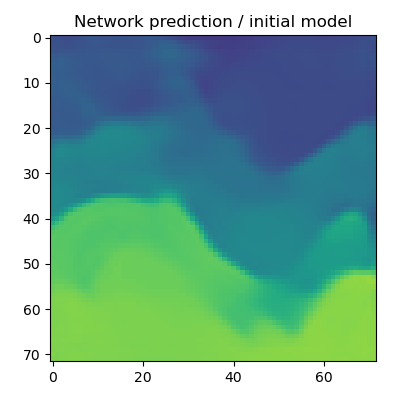}  &      \includegraphics[width=0.23\linewidth]{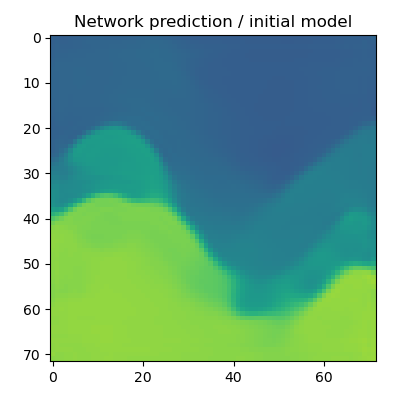} &      \includegraphics[width=0.23\linewidth]{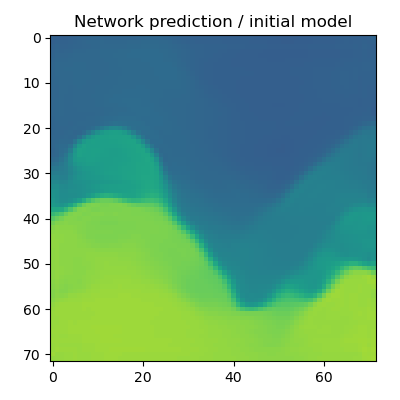} & \includegraphics[width=0.23\linewidth]{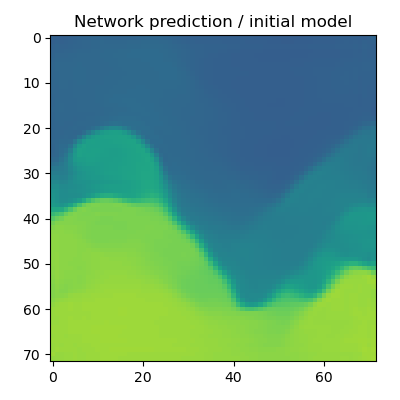} \\
    latent-space inversion \eqref{eq:LSI} & latent-space inversion \eqref{eq:LSI} & latent-space inversion \eqref{eq:LSI} & latent-space inversion \eqref{eq:LSI}\\
      \includegraphics[width=0.23\linewidth]{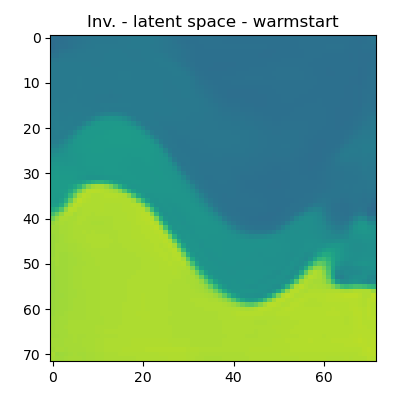}  &       \includegraphics[width=0.23\linewidth]{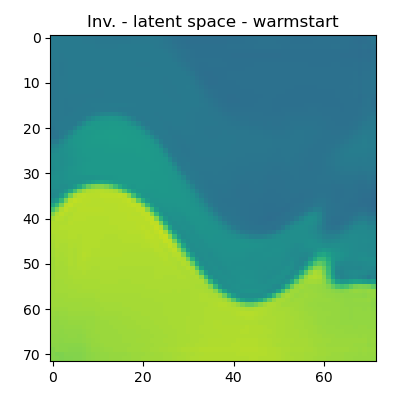} &       \includegraphics[width=0.23\linewidth]{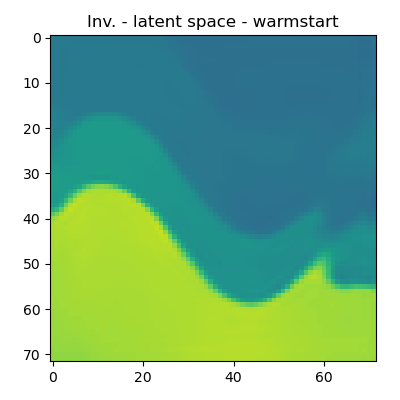} &       \includegraphics[width=0.23\linewidth]{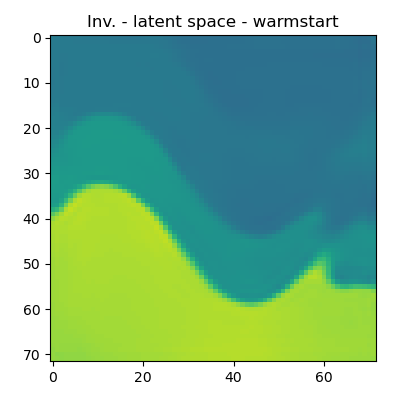} \\
      rel. error = 0.05 & rel. error= 0.05 & rel. error = 0.04 & rel. error = 0.04\\
      SSIM = 0.96 & SSIM = 0.96 & SSIM = 0.96 & SSIM = 0.96\\
       \includegraphics[width=0.23\linewidth]{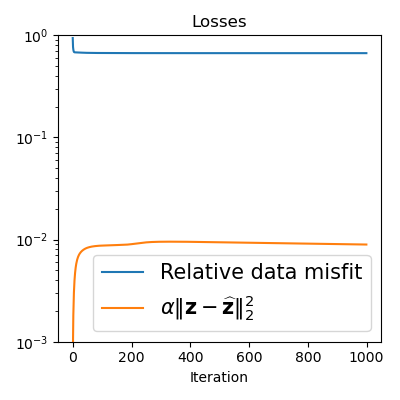}  &        \includegraphics[width=0.23\linewidth]{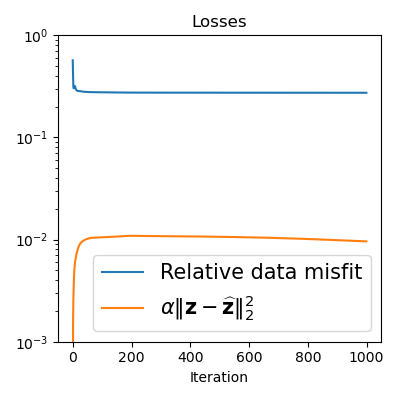} &        \includegraphics[width=0.23\linewidth]{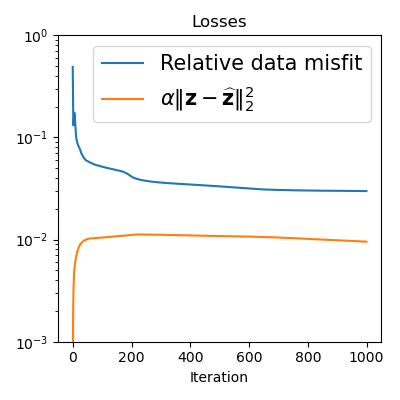} &        \includegraphics[width=0.23\linewidth]{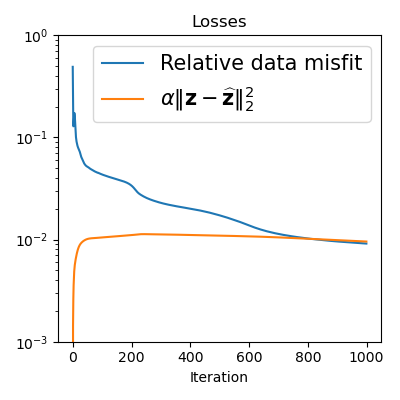}\\
      \includegraphics[width=0.23\linewidth]{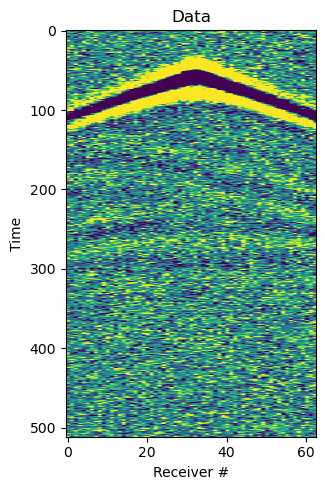}  &       \includegraphics[width=0.23\linewidth]{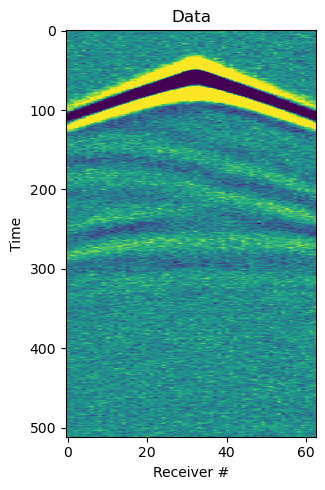} &       \includegraphics[width=0.23\linewidth]{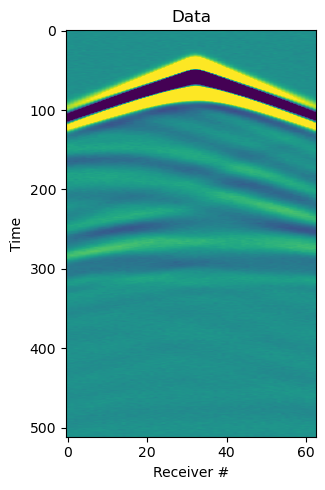} &       \includegraphics[width=0.23\linewidth]{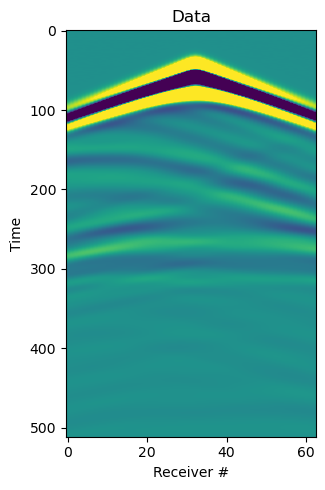} \\
    \end{tabular}
    }
    \caption{The direct estimate \eqref{eq:PAEdirect} and the refinement result from decoding the LSI inversion result \eqref{eq:LSI}, for various noise levels. The networks were trained using a fixed SNR of $30$. These results illustrate that paired autoencoders and the LSI inversion results show robustness against higher noise levels.} 
    \label{fig:FWInoisetest}
\end{figure}
         
\begin{table}
    \centering
    \begin{tabular}{c|cc}
                                                                                 & RE mean (std)  & SSIM mean (std)\\ 
                                                                                 \hline
       latent-space inversion \eqref{eq:LSI} - basic initial model               & 0.085 (0.075)  & 0.87 (0.14)\\
       latent-space inversion \eqref{eq:LSI} - warm start \eqref{eq:LSIinitial} &  0.043 (0.043) & 0.95 (0.07)\\
    \end{tabular}
    \caption{Validation set statistics for the seismic FWI problem, computed and averaged over $650$ examples in terms of relative $\ell_2$ error and structural similarity index.}
    \label{tab:PAEFWIstats}
\end{table}

\begin{figure}
    \centering
    \scalebox{0.7}{
    \begin{tabular}{cccc}
        True models & $\hat x$ \eqref{eq:LSIinitial} & latent-space inversion \eqref{eq:LSI} & latent-space inversion \eqref{eq:LSI}\\
            &  & basic initial model & warm start \eqref{eq:LSIinitial}\\
         \includegraphics[width=0.25\textwidth]{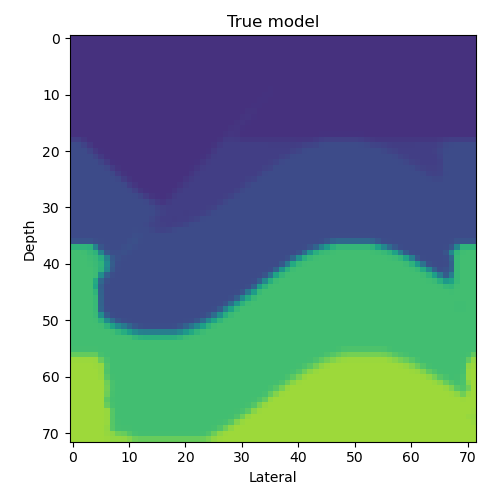} & \includegraphics[width=0.25\linewidth]{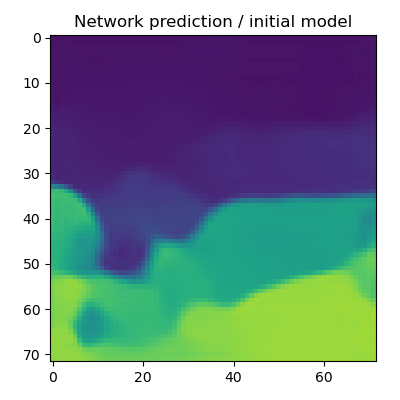} & \includegraphics[width=0.25\linewidth]{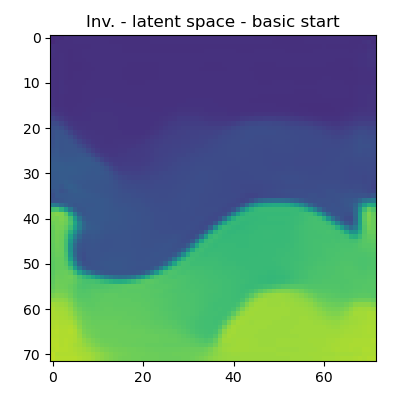} & \includegraphics[width=0.25\linewidth]{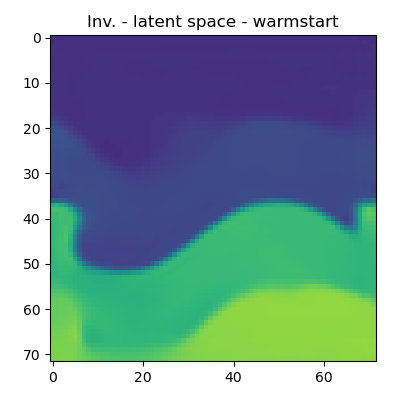} \\
         & & RE 0.04 - SSIM 0.96 & RE 0.03 - SSIM 0.97\\
         \includegraphics[width=0.25\textwidth]{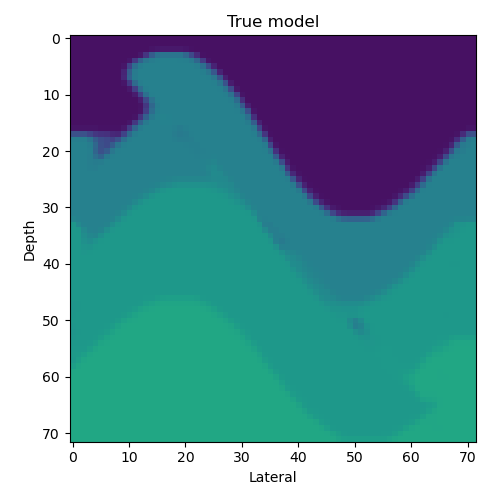} & \includegraphics[width=0.25\linewidth]{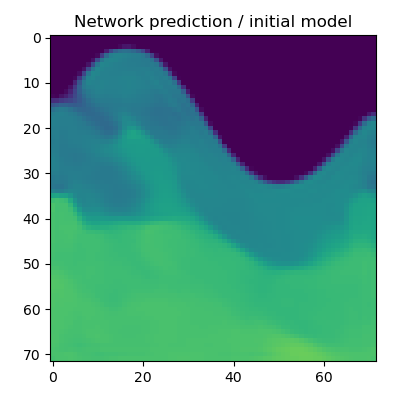} & \includegraphics[width=0.25\linewidth]{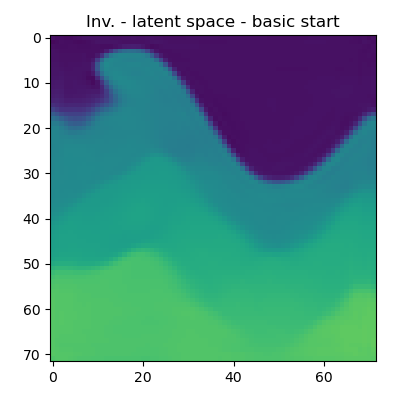} & \includegraphics[width=0.25\linewidth]{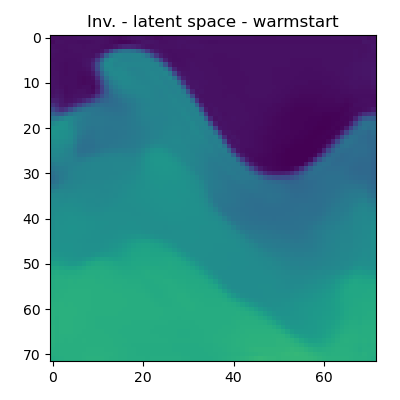} \\
         & & RE 0.09 - SSIM 0.93 & RE 0.06 - SSIM 0.90\\
    \end{tabular}
    }
    \caption{A {few validation examples} with SNR $= 10$dB seismic data. The networks were trained using $30$dB data.}
    \label{fig:SNR10validationex}
\end{figure}

\subsection{Variational Paired Autoencoders for Seismic Full-Waveform Inversion}

We revisit the previous experiment on seismic FWI, this time by training the variational paired autoencoder using Equation~\ref{eq:VPAE_training}. Unlike the regular paired autoencoder, its variational counterpart provides samples of the solution that may be analyzed using simple statistics.

Although various deep learning studies display increasingly accurate and visually near-perfect results on the datasets of the OpenFWI database \cite{deng2022openfwi}, inverse problems using seismic data remain ill-posed. Higher noise levels lead to a narrower range of usable frequencies, and more complicated velocity models or OOD data/models expose the ill-posed nature. See Figure~\ref{fig:FWInoisetest} for an example of the influence of data noise. In other words, methods that provide insight into the uncertainty of velocity models that fit given data remain important. 

Figure~\ref{fig:VPAEsamples} illustrates how the VPAE supports the objective outlined in the previous paragraph and displays eight samples as defined in \ref{VPAEsamples}. For sampling and uncertainty quantification to be practically useful, we need at least a qualitative correlation between the uncertainty and an error estimate in the predictions. Figure~\ref{fig:VPAEsummary} shows the summary computed from $100$ samples that highlights a strong correlation between the standard deviation and error in the mean estimate. See Figure~\ref{fig:FWIVPAEMeanSTD} for some more examples of the mean, standard deviation, and error computed for the validation examples. All experiments in this subsection use networks trained using data with SNR$=30$dB, while the validation data SNR$=10$dB.

This short example only scratches the surface of what is possible using the VPAE. A more refined empirical analysis involves verifying that all generated samples fit the observed data with sufficient accuracy. If samples do not fit the data, they may be discarded or serve as the initial guess for latent-space inversion (Equation~\ref{eq:LSI}) to increase the data fit. The final summary statistics are then based on samples that accurately fit the observed data.

\begin{figure}
    \centering
    \includegraphics[width=1\linewidth]{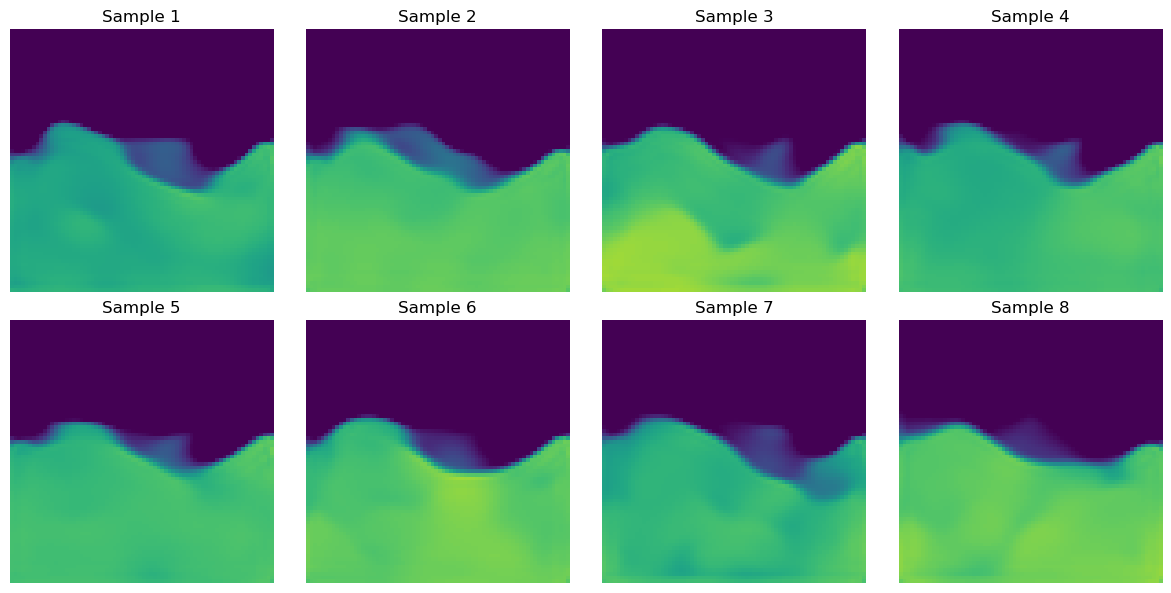}
    \caption{Eight samples of the estimated acoustic velocity model.}
    \label{fig:VPAEsamples}
\end{figure}

\begin{figure}
    \centering
    \includegraphics[width=1\linewidth]{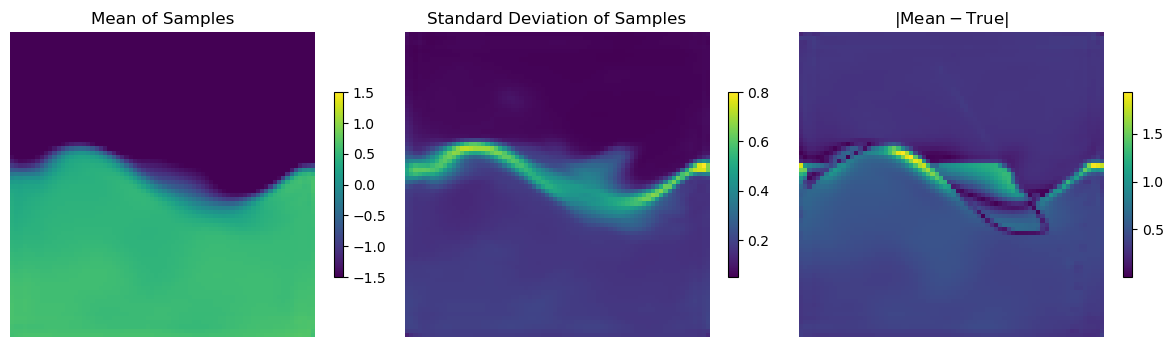}
    \caption{The mean and standard deviation computed using $100$ samples. The figure also shows the error in the mean estimate, which correlated well with the estimated standard deviation.}
    \label{fig:VPAEsummary}
\end{figure}

\begin{figure}
    \centering
    \includegraphics[width=0.65\linewidth]{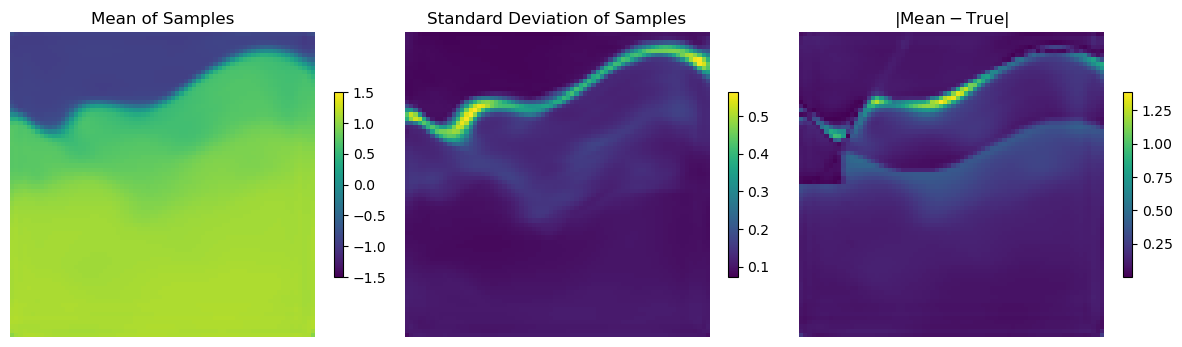}\\
    \includegraphics[width=0.65\linewidth]{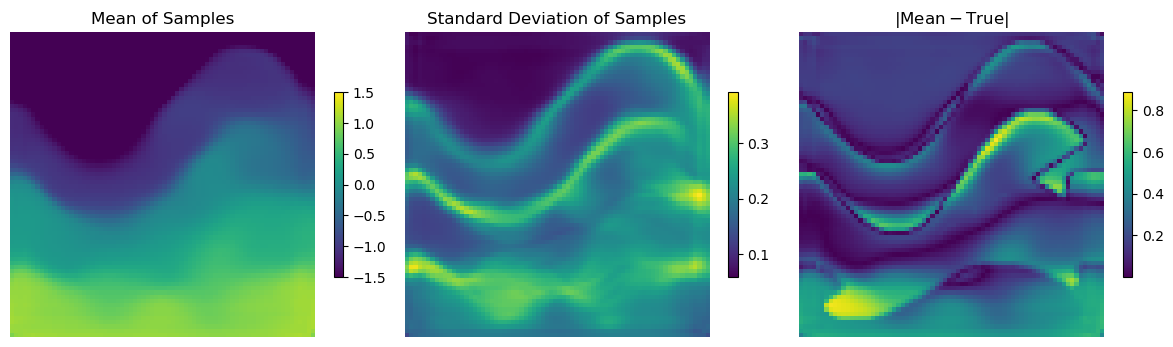}
    \includegraphics[width=0.65\linewidth]{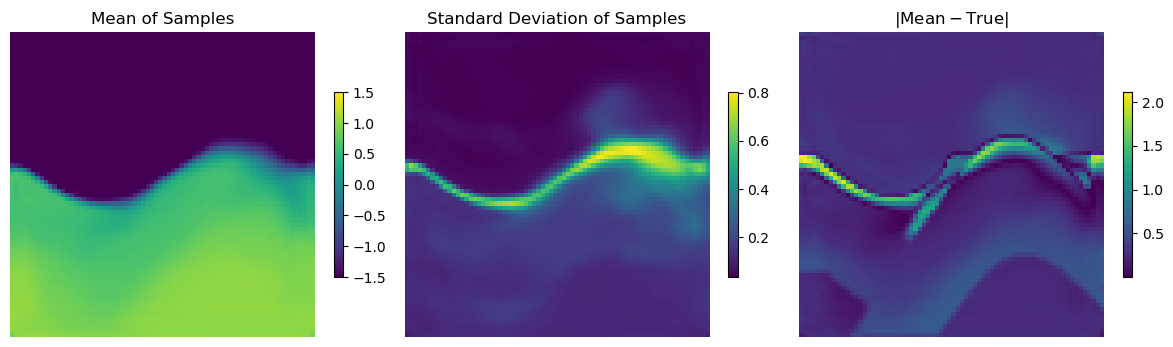}
    \caption{For three validation models in the seismic FWI problem, the figure displays the sample mean, standard deviation, and the difference between the mean and true model.}
    \label{fig:FWIVPAEMeanSTD}
\end{figure}

\section{Discussion} \label{sec:discuss}

Here, we presented a comprehensive overview of the paired autoencoder framework as a promising data-driven methodology for tackling a variety of inverse problems in scientific computing. By effectively learning the mappings between latent spaces of both the data and the quantities of interest, this approach offers a powerful surrogate model capable of approximating both forward and inverse processes. The presented numerical experiments focusing on image restoration and seismic inversion demonstrate the practical advantages and potential of paired autoencoders in yielding high-quality reconstructions. The framework’s ability to leverage both data-driven and model-based strengths opens up new avenues for addressing complex inverse problems, particularly in scenarios where traditional methods face limitations due to scale, ill-posedness, or computational cost. Training and direct inference do not require the forward-modeling operator; only latent-space inversion does. The underlying challenge of all data-driven approaches is observed--poor generalization to OOD samples.  Among the data-driven methods the paired autoencoder approach is advantageous in that it provides a way to predict performance for OOD samples through comparison of the quantities discussed.  This is advantageous in application use cases where the data is not well understood. Including a latent space optimization allows for model conform inversion as demonstrated in the seismic application. The inclusion of a variational paired autoencoder in this study allowed for uncertainty quantification of the quantity of interest while simultaneously facilitating a more well-defined latent space. 

While this paired autoencoder framework is in its infancy, many directions, even beyond inverse problems, may be considered, including adjoint estimation, latent space representation, dynamical systems dynamics, and digital twins in the latent space. Current work delves deeper into exploring the robustness of paired autoencoders to OOD data, further refining the likelihood-free inference techniques, uncertainty estimation, and investigating their applicability to a wider range of challenging scientific and engineering inverse problems.

\paragraph*{Acknowledgment.} 
The National Science Foundation partially supported this research through the CDS\&E program of the Division of Mathematical Sciences under grant number DMS-2152661.

\section{Neural Network Designs} \label{appendix}

\subsection{Image Restoration}\label{appendix:imagerestoration}
Identical encoder/decoder architectures were used for both autoencoders in the image restoration problems using $28 \times 28$ input images. We implement a 6-layer 2D-CNN for our encoder with $3\times3$ kernels while maintaining a padding of $1$.  The first layer increases the number of channels to $16$ and subsequent layers double the number of channels, while halving the spatial dimensions.  Here, the spatial downsampling occurs through average pooling with a kernel size and stride of 2. The decoder is implemented using a similar architecture, starting with $128$ channels which are mapped to a $128 \times 8 \times 8$ tensor via a linear mapping.  From here, each block doubles the spatial dimension at each decoder layer while halving the number of channels.  Since we upsample by $2\times$ at each decoder layer, an additional interpolation is required and is implemented using a nearest-neighbors interpolation algorithm. The full decoder block sequence is given as the following composition of layers
\begin{align}\label{eq:decoderlayer}
    \bfX_{i} \rightarrow {\rm Upsampling\; by \; }2\times \rightarrow {\rm Conv2D} \rightarrow {\rm Batch \; Norm} \rightarrow {\rm SiLU} \rightarrow {\rm Conv2D} \rightarrow \bfX_{i + 1}.
\end{align}
Here, the first convolution layer does not augment the dimensions of the image, where the second one reduces the number of channels by half.  Both latent space mappings are simple trainable linear maps, capped with a sigmoid function. The encoder and decoder architectures are summarized in Table~\ref{table:image}.  

\begin{table}

    \centering
    \begin{tabular}{lcl}
        \multicolumn{3}{c}{\textbf{Image Restoration Encoder}}\\
        \textbf{Layer} & \textbf{Output size} & \textbf{Type}\\
        \hline
         input & $1 \times 28 \times 28$ & nChannels $\times$ length $\times$ width \\
         1 & $16 \times 28 \times 28$ & Conv2D \\
         2 & $32 \times 14 \times 14$ & Conv2D \\
         3 & $64 \times 7 \times 7$ & Conv2D\\
         4 & 3136 & Flatten \\
         5 & 128 & Linear\\
         6 & 128 & ReLU \\
        \hline\\
        
         \hline\hline\\
     \end{tabular}

        \begin{tabular}{lcl}
        \multicolumn{3}{c}{\textbf{Image Restoration Decoder}}\\
        \textbf{Layer} & \textbf{Output size} & \textbf{Type}\\
        \hline
         input & $128 \times1 \times 1$ & nLatent $\times$ length $\times$ width \\

        1 & $128 \times 8 \times 8$ & Linear \\
         2-7 & $64 \times 16\times 16$ & Equation~\ref{eq:decoderlayer} \\
         8-12 & $32 \times 32 \times 32$ & Equation~\ref{eq:decoderlayer}\\
         13-18 & $16 \times 64 \times 64$ & Equation~\ref{eq:decoderlayer}\\

         19 & $1 \times 28 \times 28$ & Conv2D\\
         20 & $1 \times 28 \times 28$ & Sigmoid\\
        \hline\\
        
         \hline\hline\\
     \end{tabular}
     \caption{Image Restoration encoder and decoder descriptions.}
     \label{table:image}
 \end{table}

\subsection{Variational Latent Mapping for Image Restoration}\label{sect:ProbLatNetwork}

The encoder/decoder architecture used for the probabilistic latent mapping experiment are identical to Appendix~\ref{appendix:imagerestoration}.  The probabilistic encoder and decoder used for $M^{\dagger}$ is summarized in Table~\ref{table:prob}.  Here we note that the encoder is non-compressive--the value sampled from the probabilistic encoder has dimension $128$ which is equivalent to the dimension of the data latent space.

\begin{table}

    \centering

         \begin{tabular}{lcl}
        \multicolumn{3}{c}{\textbf{Image Restoration {Latent Mapping} Encoder}}\\
        \textbf{Layer} & \textbf{Output size} & \textbf{Type}\\
        \hline
         input & $128$ & nLatent\\
         1 & $128$  & Linear (hidden dimension $= 128$) \\
         2 & $2 \times 128$ & Linear \\
         3 & $2 \times 128$ & ReLU\\
    
        \hline\\
        
         \hline\hline\\
     \end{tabular}

         \begin{tabular}{lcl}
        \multicolumn{3}{c}{\textbf{Image Restoration {Latent Mapping}  Decoder}}\\
        \textbf{Layer} & \textbf{Output size} & \textbf{Type}\\
        \hline
         input & $128$ & nLatent\\
         1 & $128$  & Linear (hidden dimension $= 128$) \\
         2 & $128$ & ReLU \\
         3 & $128$ & Linear \\
    
        \hline\\
        
         \hline\hline\\
     \end{tabular}
     \caption{Encoder/Decoder architectures used for the probabilistic latent space experiment.  Note that the output of the encoder is $2 \times 128$, since a mean and log-variance are required to sample from the latent distribution.}
     \label{table:prob}
 \end{table}

\subsection{Seismic FWI} 
The encoder and decoder for the seismic data and velocity model are both an instance of a multi-level ResNet \cite{he2016deep}. Because the seismic data (nr. sources $\times$ nr. receivers $\times$ nr. time samples) and velocity model ($n_z \times n_x$) are of different sizes and dimensions, we design different networks for each. 

The ResNet block is the core of all our networks, and is given by 
\begin{equation}\label{eq:resnetblock}
    \bfX_{i+1} = \bfX_i - h \bfK_1 \sigma(N(\bfK_2 \bfX_i)).
\end{equation}
Here, $\bfX_{i+1}$ is the network state at layer $i$, $N(\cdot)$ is the batch-normalization function, $\sigma(\cdot)$ is the SiLU nonlinearity, $h$ is the artificial time-step (selected as $0.2$) and finally, $\bfK_1$ and $\bfK_2$ are two learnable linear operators (block-convolutional matrices in our case).

The encoder and decoders consist of multiple levels (resolutions). We design one such level in the encoder as

\begin{equation}\label{eq:resnetblockfull}
    \bfX_{i+1} = \bfK_{cc} \bfX_i \rightarrow 3\times \text{ResNet block (Equation~\ref{eq:resnetblock})} \rightarrow \bfX_{i+1} = P(\bfX_i),
\end{equation}
where the number of channels of the input state $\bfX_{i}$ gets modified by the convolutional operator $\bfK_{cc}$, followed by three ResNet blocks, and ending with a pooling (downsampling) operation $P(\cdot)$. The design of the decoder adheres to the reverse of the encoder design, i.e.,

\begin{equation}\label{eq:resnetblockfull2}
    \bfX_{i+1} = U(\bfX_i) \rightarrow \bfX_{i+1} = \bfK_{cc} \bfX_i \rightarrow 3\times \text{ResNet block (Equation~\ref{eq:resnetblock})},
\end{equation}
where $U(\cdot)$ interpolates/upsamples the state. The pooling and upsampling operations may act on both the temporal/spatial or spatial/spatial directions, or on just the temporal direction.

The final latent vector arises from vectorizing the network state and multiplication with matrix $\bfA$. See Table~\ref{tab:seismicdataencoderdecoder} and Table~\ref{tab:seismicmodelencoderdecoder} for the data encoder and decoder details in terms of dimensions, where the matrix $\TA$ is always a linear operator of appropriate dimensions.

The latent mappings $M$ and $M^\dagger$ are learnable matrices of size $1280 \times 1280$.

\begin{table}
    \centering
    \begin{tabular}{lcl}
        \multicolumn{3}{c}{\textbf{Seismic Data Encoder}}\\
        \textbf{Layer} & \textbf{Feature size} & \textbf{Type}\\
        \hline
         input & $30 \times 72 \times 512$ & channels(sources) $\times$ receivers $\times$ time\\
         1-3 & $32 \times 72 \times 512$ & Equation~\ref{eq:resnetblockfull} \\
         4-6 & $83 \times 72 \times 256$ & Equation~\ref{eq:resnetblockfull} \\
         7-9 & $134 \times 72 \times 128$ & Equation~\ref{eq:resnetblockfull} \\
         10-12 & $185 \times 72 \times 64$ & Equation~\ref{eq:resnetblockfull} \\
         13-15 & $256 \times 36 \times 32$ & Equation~\ref{eq:resnetblockfull} \\
         16-18 & $256 \times 18 \times 16$ & Equation~\ref{eq:resnetblockfull} $\times 3$\\
         19 & $1280$ & $\bfX_{i+1} = \bfA\vec{\bfX_i}$\\
        \hline\\
        
         \hline\hline\\

        \multicolumn{3}{c}{\textbf{Seismic Data Decoder}}\\
        \textbf{Layer} & \textbf{Feature size} & \textbf{Type}\\
        \hline
         input & $1280$ & Latent variables\\
         1 & $128 \times 9 \times 18$ & $\bfX_{i+1} = \bfA \bfX_i \rightarrow$ reshape\\
         2-4 & $256 \times 18 \times 16$ & Equation~\ref{eq:resnetblockfull} $\times 3$\\
         5-7 & $256 \times 36 \times 32$ & Equation~\ref{eq:resnetblockfull} \\
         8-10 & $185 \times 72 \times 64$ & Equation~\ref{eq:resnetblockfull} \\
         11-13 & $134 \times 72 \times 128$ & Equation~\ref{eq:resnetblockfull} \\
         14-16 & $83 \times 72 \times 256$ & Equation~\ref{eq:resnetblockfull} \\
         17-19 & $32 \times 72 \times 512$ & Equation~\ref{eq:resnetblockfull} \\
         \hline
    \end{tabular}
    \caption{Network for the seismic data encoder $e_y$ and decoder $d_y$.}
    \label{tab:seismicdataencoderdecoder}
\end{table}

\begin{table}
    \centering
    \begin{tabular}{lcl}
        \multicolumn{3}{c}{\textbf{Seismic Model Encoder}}\\
        \textbf{Layer} & \textbf{Feature size} & \textbf{Type}\\
        \hline
         input & $1 \times 72 \times 72$ & channels(sources) $\times$ receivers $\times$ time\\
         1-3 & $16 \times 72 \times 72$ & Equation~\ref{eq:resnetblockfull} \\
         4-6 & $32 \times 36 \times 36$ & Equation~\ref{eq:resnetblockfull} \\
         7-9 & $64 \times 18 \times 18$ & Equation~\ref{eq:resnetblockfull} \\
         10-12 & $64 \times 18 \times 18$ & Equation~\ref{eq:resnetblock} $\times 3$\\
         13 & $30 \times 18 \times 18$ & $ \bfX_{i+1} = \bfK_{cc} \bfX_i$\\
         14 & $1280$ & $\bfX_{i+1} = \bfA\vec{\bfX_i}$\\
        \hline\\
        
         \hline\hline\\

        \multicolumn{3}{c}{\textbf{Seismic Model Decoder}}\\
        \textbf{Layer} & \textbf{Feature size} & \textbf{Type}\\
        \hline
         input & $1280$ & Latent variables\\
         1 & $30 \times 18 \times 18$ & $\bfX_{i+1} = \bfA \bfX_i \rightarrow$ reshape\\
         2 & $64 \times 18 \times 18$ & $ \bfX_{i+1} = \bfK_{cc} \bfX_i$\\
         3-5 & $64 \times 18 \times 18$ & Equation~\ref{eq:resnetblock} $\times 3$\\
         6-8 & $64 \times 18 \times 18$ & Equation~\ref{eq:resnetblockfull2}\\
         9-11 & $32 \times 36 \times 36$ & Equation~\ref{eq:resnetblockfull2}\\
         12-14 & $16 \times 72 \times 72$ & Equation~\ref{eq:resnetblockfull2} \\
         15 & $1 \times 72 \times 72$ & $ \bfX_{i+1} = \bfK_{cc} \bfX_i$\\
         \hline
    \end{tabular}
    \caption{Network for the seismic velocity model encoder $e_x$ and decoder $d_x$.}
    \label{tab:seismicmodelencoderdecoder}
\end{table}

\pagebreak

\bibliographystyle{plainnat}
\bibliography{references}

\end{document}